\documentclass[10pt,twocolumn,letterpaper]{article}

\usepackage{iccv}
\usepackage{times}
\usepackage{epsfig}
\usepackage{graphicx}

\usepackage{amsmath, amsthm, amssymb, mathabx}
\usepackage{textcomp}
\usepackage{stmaryrd}
\SetSymbolFont{stmry}{bold}{U}{stmry}{m}{n}
\usepackage{upgreek}
\usepackage{bm}
\usepackage{cases}
\usepackage{mathtools}
\usepackage{arydshln}
\usepackage{times}
\usepackage{lipsum}
\usepackage{comment}
\usepackage{subfig}
\usepackage{appendix}

\usepackage{cite}

\usepackage{algpseudocode,algorithm,algorithmicx}

\usepackage{multirow}
\usepackage{rotating}
\usepackage{booktabs}

\usepackage{enumitem}
\usepackage[olditem,oldenum]{paralist}

\usepackage{alltt}
\usepackage{listings}

\usepackage{mysymbols}

\usepackage{url}
\usepackage{comment}
\usepackage{color}
\usepackage[dvipsnames]{xcolor}
\usepackage{afterpage}
\usepackage{pdfpages}
\usepackage{framed}
\usepackage{fancybox}
\usepackage{cuted}
\usepackage{nicefrac}
\usepackage{quoting}
\quotingsetup{vskip=0pt}
\usepackage{epigraph}

\usepackage{xspace}
\usepackage{subfig}
\usepackage{float}
\usepackage{contour}
\usepackage{xkeyval}

\usepackage[pagebackref=true,breaklinks=true,letterpaper=true,colorlinks,bookmarks=false]{hyperref}

\iccvfinalcopy %

\def\iccvPaperID{320} %

\ificcvfinal\fi
\begin{document}

\newcommand{\wh}{\color{white}} %
\contourlength{.4pt}
\newcommand{\figlabel}[1]{\sffamily\bfseries\wh\scriptsize\contour{black}{#1}} %
\makeatletter
\newlength{\sfp@hseplen}\newlength{\sfp@vseplen}
\define@cmdkey{subfigpos}[sfp@]{vsep}[0.6\baselineskip]{\setlength{\sfp@vseplen}{\sfp@vsep}}%
\define@cmdkey{subfigpos}[sfp@]{hsep}[2.5pt]{\setlength{\sfp@hseplen}{\sfp@hsep}}%
\newcommand{\subfigimg}[3][,]{%
  \setkeys{Gin,subfigpos}{vsep,hsep,#1}%
  \setbox1=\hbox{\includegraphics{#3}}%
  \leavevmode\rlap{\usebox1}%
  \rlap{\hspace*{\sfp@hsep}\raisebox{\dimexpr\ht1-6pt}{\figlabel{#2}}}%
  \phantom{\usebox1}%
}
\makeatother

\newcommand{\devi}[1]{\textcolor{red}{#1}}
\newcommand{\dhruv}[1]{\textcolor{red}{DB: #1}}
\newcommand{\jianwei}[1]{\textcolor{blue}{#1}}
\newcommand{\mingze}[1]{\textcolor{brown}{MX: #1}}
\newcommand{\xinlei}[1]{\textcolor{orange}{XC: #1}}
\newcommand{\ren}[1]{\textcolor{cyan}{ZR: #1}}
\newcommand{\checkthis}[1]{\textcolor{red}{CheckThis: #1}}

\title{Embodied Visual Recognition}
\author{
Jianwei Yang$^{1 \star}$,
Zhile Ren$^{1 \star}$, 
Mingze Xu$^2$, \\
Xinlei Chen$^3$, 
David Crandall$^2$, 
Devi Parikh$^{1,3}$ 
Dhruv Batra$^{1,3}$ \\
$^1$Georgia Institute of Technology, $^2$Indiana University, $^3$Facebook AI Research.\\
\small{\url{https://www.cc.gatech.edu/~jyang375/evr.html}}
}

\maketitle

\renewcommand*{\thefootnote}{$\star$}
\setcounter{footnote}{1}
\footnotetext{The first two authors contributed equally.}
\renewcommand*{\thefootnote}{\arabic{footnote}}
\setcounter{footnote}{0}

\begin{abstract}
Passive visual systems typically fail to recognize objects in the amodal setting where they are heavily occluded. In contrast, humans and other embodied agents have the ability to move in the environment, and actively control the viewing angle to better understand object shapes and semantics. In this work, we introduce the task of Embodied Visual Recognition (EVR): An agent is instantiated in a 3D environment close to an occluded target object, and is free to move in the environment to perform object classification, amodal object localization, and amodal object segmentation. To address this, we develop a new model called Embodied Mask R-CNN, for agents to learn to move strategically to improve their visual recognition abilities. We conduct experiments using the House3D environment. Experimental results show that: 1) agents with embodiment (movement) achieve better visual recognition performance than passive ones; 2) in order to improve visual recognition abilities, agents can learn strategical moving paths that are different from shortest paths.
\end{abstract}

\section{Introduction}
\label{sec:intro}

\setlength{\epigraphwidth}{0.7\columnwidth}
\renewcommand{\textflush}{flushepinormal}

\begin{figure}[t]
	\centering
	\includegraphics[width=1.0\linewidth]{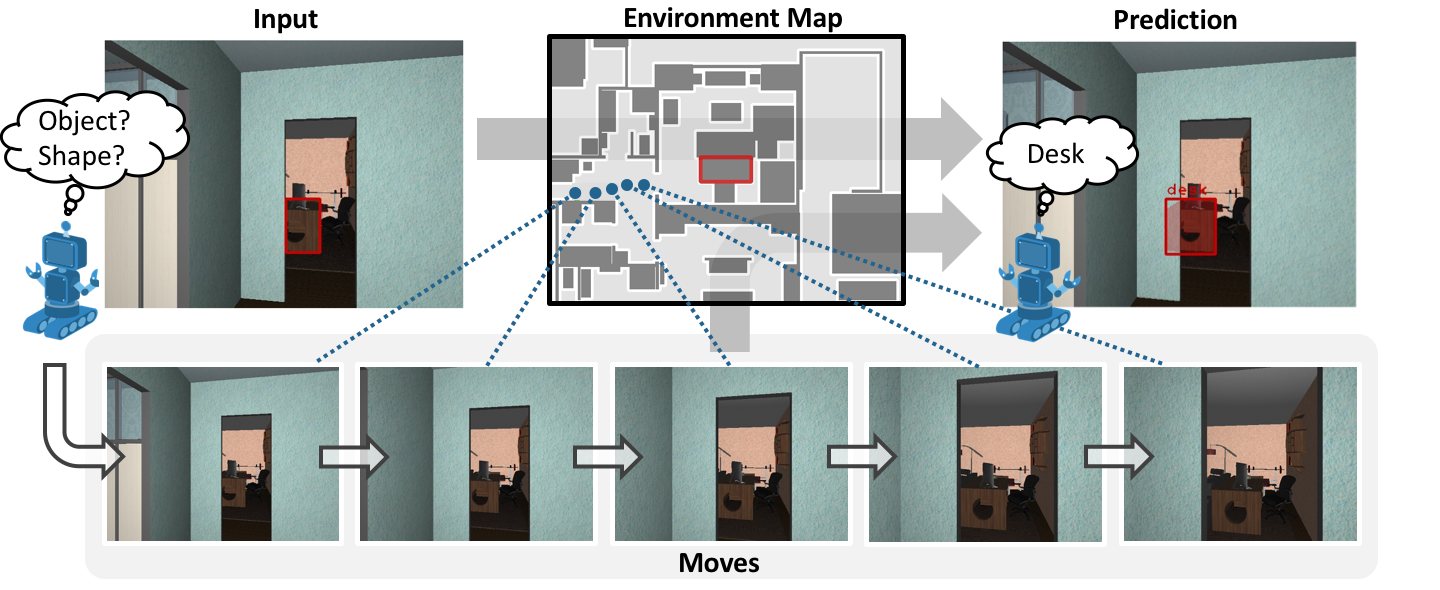}
	\vspace{-15pt}	
    \caption{The task of Embodied Visual Recognition: An agent is spawned close to an occluded target object in a 3D environment, and asked for visual recognition, \ie, predicting class label, amodal bounding box and amodal mask of the target object. The agent is free to move around to aggregate information for better visual recognition.}
	\label{fig:teaser}
		\vspace{-5pt}	
\end{figure}

Recently, visual recognition tasks such as image classification~\cite{krizhevsky2012imagenet,szegedy2015going,he2016deep,hu2018senet}, object detection~\cite{girshick2014rich, girshick2015fast, ren2015faster, redmon2016you, yolov3} and semantic segmentation~\cite{long2015fully,yu2015multi,zhao2017pyramid}, have been widely studied. In addition to recognizing the object's semantics and shape for its visible part, the ability to perceive the whole of an occluded object, known as amodal perception~\cite{kanizsa1979Organization,stephen1999vision,wagemans2012century}, is also important. Take the desk (red bounding box) in the top-left of \figref{fig:teaser} as an example, the amodal predictions (top-right of \figref{fig:teaser}) can tell us about the depth ordering (\ie, desk is behind the wall), the extent and boundaries of occlusions, and even estimations of physical dimensions~\cite{kar2015amodal}. More fundamentally, it helps agents to understand object permanence, that is, objects have extents and do not cease to exist when they are occluded~\cite{baillargeon1985object}.

Recently, the dominant paradigm for object recognition and amodal perception has been based on single image. Though leveraging the advances of deep learning, visual systems still fail to recognize object and its shape from single 2D image in the presence of heavy occlusions. Consider amodal perception. Existing works ask the model to implicitly learn the 3D shape of the object \emph{and} the projection of that shape back into the image~\cite{li2016amodal, zhu2017semantic,ehsani2018segan}. This is an entangled task, and deep models are thus prone to over-fit to subtle biases in the dataset~\cite{geirhos2018imagenet} (\eg learning that beds always extend leftwards into the frame).

Remarkably, humans have the visual recognition ability to infer both semantics and shape for the occluded objects from a single image. On the other hand, humans also have the ability to derive strategical moves to gather more information from new viewpoints to further help the visual recognition. A recent study in \cite{bambach2018toddler} shows that toddlers are capable of actively diverting viewpoints to learn about objects, even when they are only 4\,--\,7 months old.

Inspired by human vision, the key thesis of our work is that in addition to \emph{learning to hallucinate}, agents should \emph{learn to move} as well. 
As shown in \figref{fig:teaser}, to recognize the category and whole shape of target object indicated by the red bounding box, agents should learn to actively move toward the target object to unveil the occlude region behind the wall for better recognition.

In this paper, we introduce a new task called \emph{Embodied Visual Recognition} (EVR) where agents actively move in a 3D environment for visual recognition of a target object. We are aimed at systemically studying whether and how embodiment (movement) helps visual recognition. Though it is a general paradigm, we highlight three design choices for the EVR task in this paper:

\xhdr{Three sub-tasks.} In EVR, we aim to recover both semantics and shape for the target object. It consists of three sub-tasks: object recognition, 2D amodal perception (amodal localization and amodal segmentation). With these three sub-tasks, we provide a new test bed for vision systems.

\xhdr{Single target object.} When spawned in a 3D environment, the agent may see multiple objects in the field-of-view. We specify one instance as the target, and denote it using a bounding box encompassing its \emph{visible} region. The agent's goal then is to move to perceive this single target object.

\xhdr{Predict for the first frame.} The agent performs visual recognition for the target object observed at the spawning point. If the agent does not move, EVR degrades to a passive visual recognition. Both passive and embodied algorithms are trained using the same amount of supervisions and evaluated on the same set of images. As a result, we can create a fair benchmark to evaluate different algorithms.

\begin{figure}[!t]
	\centering
	\includegraphics[width=0.98\linewidth]{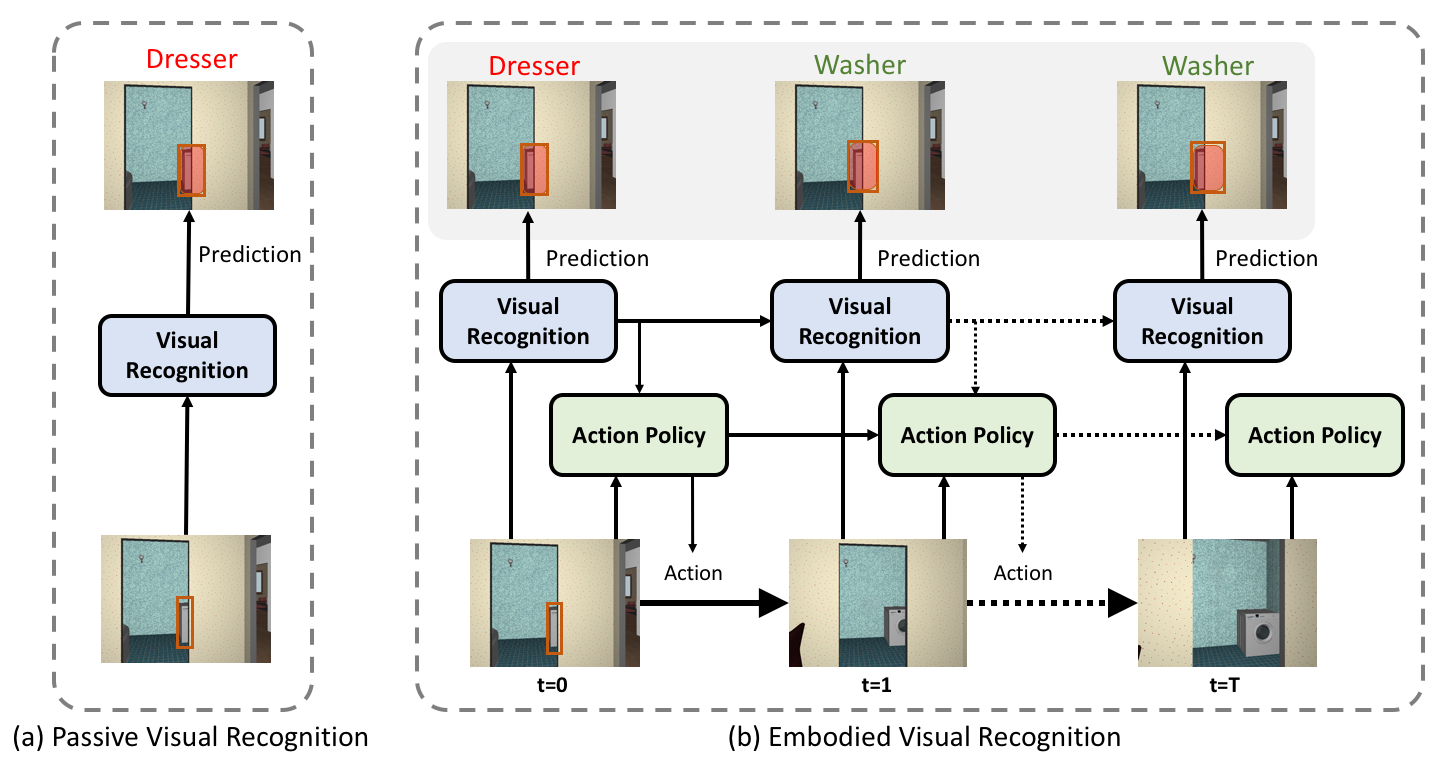}
	\vspace{-5pt}
    \caption{
        The proposed general pipeline for Embodied Visual Recognition task. To perform visual recognition (object recognition and amodal perception) on the occluded object, the agent \emph{learns to move} (right), rather than standing still and \emph{hallucinating} (left). The visual recognition module focuses on predicting the object class, amodal bounding box and masks for the first frame. The policy module proposes the next move for the agent to acquire useful information about the object.
    }
	\label{fig:framework}
\end{figure}

Based on the above choices, we propose a general pipeline for EVR as shown in \figref{fig:framework}. Compared with the passive visual recognition model (\figref{fig:framework}a), the embodied agent (\figref{fig:framework}b) will follow the proposed action from the policy module to move in the environment, and make the predictions on the target object using the visual recognition module. This pipeline introduces several interesting problems: 1) Due to agent's movement, the appearances of observed scene \emph{and} target object change in each step. How should information be aggregated from future frames to the first frame for visual recognition? 
2) There is no ``expert'' that can tell the agent how to move in order to improve its visual recognition performance. How to effectively propose a strategic move without any supervision? 
3) In this task, the perception module and action policy are both learned from scratch. Considering the performance of each heavily relies on the competence of the other, how to design proper training regime is also an open question.

To address the above questions, we propose a new model called \emph{Embodied Mask R-CNN}. The perception module extends Mask R-CNN~\cite{he2017mask} by adding a recurrent network to aggregate temporal features. The policy module takes the current observation and features from the past frames to predict the action. We use a staged training scheme to train these two modules effectively.

\xhdr{Contributions.} The main contributions of this paper are: 
\begin{compactitem}

\item We introduce a new task, Embodied Visual Recognition, where an agent can move in a 3D environment to perform 2D object recognition and amodal perception, including amodal localization and segmentation.
\item We build a new dataset for EVR. Using the House3D simulator~\cite{wu2018building} on SUNCG~\cite{song2016ssc}, we collect viewpoints for agents so that the target object is partially visible. We also provide precise ground-truth annotations of object classes, amodal bounding boxes and masks.

\item We present a general pipeline for EVR and propose a new model, Embodied Mask R-CNN, to learn to move for visual recognition. In this model, the visual recognition and policy module make predictions at each step, and aim to improve the visual recognition performance on the target object in the first frame.

\item We evaluate both passive and embodied vision recognition systems, and demonstrate that agents with movements consistently outperform passive ones. Moreover, the learned moves are more effective in improving visual recognition performance, as opposed to random or shortest-path moves.

\item We observe the emergence of interesting agent behaviors: the learned moves are different from shortest-path moves and generalize well to unseen environments (\ie, new houses and new instances of objects). 

\end{compactitem}

\section{Related Work}

\vspace{-5pt}
\xhdr{Visual Recognition}. 
Building visual recognition systems is one of the long-term goals of our community. Training on large-scale datasets~\cite{ImageNet,zhou2017places,lin2014microsoft}, we have recently witnessed the versatility and effectiveness of deep neural networks for many tasks, including image classification \cite{krizhevsky2012imagenet,szegedy2015going,he2016deep,hu2018senet}, object detection \cite{girshick2014rich, girshick2015fast, ren2015faster, redmon2016you, yolov3}, semantic segmentation~\cite{long2015fully,yu2015multi,zhao2017pyramid}, instance segmentation \cite{he2017mask} \etc. Extending its successful story, similar pipelines have been applied to \emph{amodal perception} as well, notably for amodal segmentation~\cite{li2016amodal,zhu2017semantic,follmann2018learning, ehsani2018segan}.

Despite these advances, visual systems still fail to recognize objects from single 2D images in the presence of significant occlusion and unusual poses. Some work has attempted to overcome this by aggregating multiple frames or views~\cite{novotny2017learning, bao2018cnn, xiao2018video,su2015multi}, other leveraging CAD models~\cite{Aubry14,gupta2015aligning,bansal2016marr,izadinia2017im2cad,sun2018pix3d}. However, a diverse set of viewpoints or CAD model is not always available \emph{a priori}, and unlikely to hold in practice. We would like to build the capability of agents to move around and change viewing angle in order to perceive. This is the goal of \emph{active vision}. 

\xhdr{Active Vision}. Active vision has a long history of research~\cite{bajcsy1988active,aloimonos1988active,wilkes1992active}, and also has connections to developmental psychology~\cite{bambach2018toddler}. Recent work learns active strategy for object recognition~\cite{denzler2002information, jayaraman2015learning, malmir2015deep, kragic2005vision, johns2016pairwise, jayaraman2018end}, object localization/detection~\cite{gonzalez2015active,caicedo2015active,mathe2016reinforcement}, object manipulation~\cite{Cheng2018visual} and instance segmentation~\cite{pathak2018learning}. However, all of them assume a constrained scenario where either a single image is provided or the target object is localized in different views. Moreover, the agent is not embodied in a 3D environment, and thus no movement is required. Ammirato~\etal~\cite{ammirato2017dataset} built a realistic dataset for active object instance classification~\cite{han2019active}. Though involving movement, they have a similar setting to the aforementioned works, \ie, searching for a good viewpoint for instance classification by assuming the bounding boxes of the target object are known during the whole movement. In contrast, the formulation of EVR is more realistic and challenging -- we allow an embodied agent in a 3D simulator to actively move and perform visual recognition. The agent is required to have both a smart moving strategy to control what visual input to receive, and a good visual recognition system to aggregate temporal information from multiple viewpoints. 

\xhdr{Embodiment}. Recently, a number of 3D simulators have been introduced to model virtual embodiment. Several of them are based on real-world environments~\cite{mattersim,xiazamirhe2018gibsonenv,ammirato2017dataset} for tasks such as robot navigation \cite{mattersim,ye2018active} and scene understanding~\cite{2017armeni2d3d}. Other simulators have been built for synthetic environments~\cite{brodeur2017home,savva2017minos,ai2thor}, such as House3D~\cite{wu2018building}. They come in handy with accurate labels for 3D objects and programmable interfaces for building various tasks, such as visual navigation \cite{zhu2017target} and embodied question answering~\cite{embodiedqa,eqa_modular,gordon2018iqa}. EVR is a new exploration in the task space on these environments: Unlike visual navigation, where the goal is to find objects or locations, our task assumes the target object is already (partially) observed at the beginning; Unlike question answering~\cite{embodiedqa,eqa_modular,gordon2018iqa}, we only focus on visual recognition and is arguably better suited for benchmarking progress and diagnosing vision systems.

\section{Dataset for EVR}

\vspace{-5pt} \xhdr{Environment.} Although EVR can be set up on any simulation environments~\cite{mattersim,savva2017minos,ai2thor}, in this paper we use House3D~\cite{wu2018building} as a demonstration. House3D is an open-sourced simulator built on top of SUNCG~\cite{song2016ssc}, which contains objects from 80 distinct categories. Similar to the EQA dataset~\cite{embodiedqa}, we filter out atypical 3D rooms in House3D that are either too big or have multiple levels, resulting in 550 houses in total. A detailed list of houses can be found in the Appendix. These houses are split to 400, 50, 100 for training, validation and test, respectively.

\begin{figure}[t]
	\centering
	\includegraphics[width=0.95\linewidth]{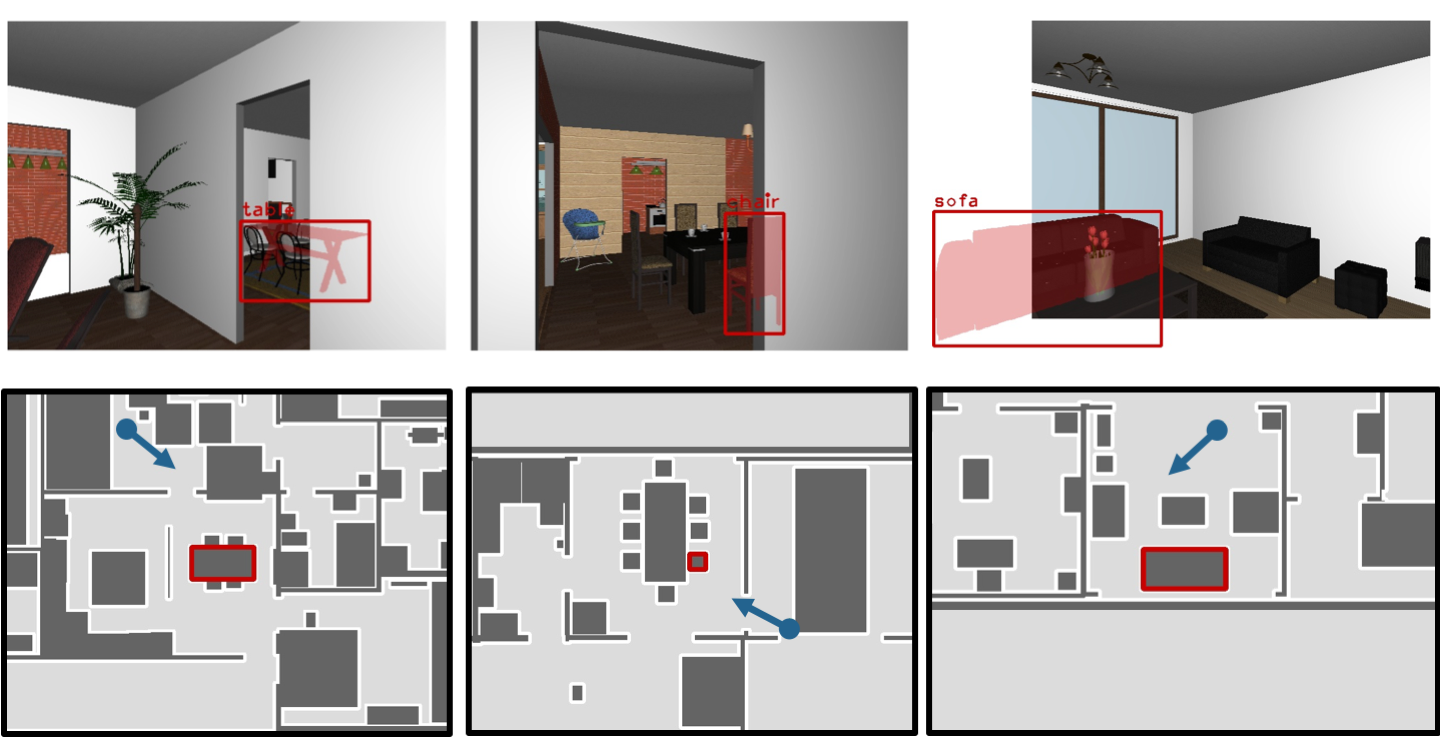}
	\vspace{-5pt}
    \caption{Example annotations in our dataset. In each column, the top row shows the ground-truth annotation for the image, and the bottom row shows the corresponding agent's location and viewpoint (blue arrow) and target object's location (red bounding box) in top-view map. From left to right, we show an easy, a hard and a partially out-of-view sample, which is not included in previous amodal datasets~\cite{li2016amodal,zhu2017semantic}.}
	\label{fig:dataset}\vspace{-5mm}	
\end{figure}

\xhdr{Rendering.} Based on the House3D simulator, we render $640{\times}800$ images, and generate ground truth annotations for object category, amodal bounding boxes and amodal masks. Previous work on amodal segmentation~\cite{li2016amodal,zhu2017semantic,ehsani2018segan} made a design decision that clips amodal object masks at image borders. We believe this undermines the definition of amodal masks and was a limitation of using static images. Our work relies on a simulator, and thus we can easily generate amodal masks that extend beyond the image borders (see the right-most example in~\figref{fig:dataset}). In practice, we extend borders of rendered images by $80$ pixels on each side (resulting in $800{\times}960$ images).

\begin{figure}[!t]
	\centering
	\includegraphics[width=0.95\linewidth]{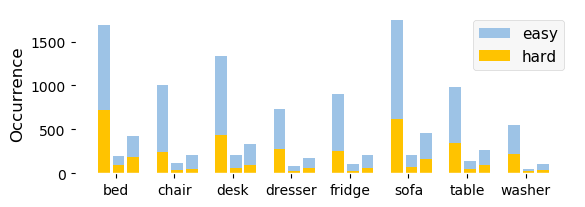}
	\vspace{-10pt}	
    \caption{Object occurrences in our dataset. For each category, the three grouped bars represent train/validation/test sets; Upper blue bars represent ``easy'' instances and bottom orange bars represent ``hard'' instances.}
	\label{fig:obj_stats}\vspace{-2mm}	
\end{figure}

\xhdr{Objects.} We select a subset of object categories that are suitable for us to study agent's understanding for occluded objects. Our selection criteria are: 1) objects should have a sufficient number of appearances in the training data, 2) objects should have relatively rigid shapes and not have deformable structures (curtains, towels, \etc), ambiguous geometries (toys, paper, \etc), or be room components (floors, ceilings, \etc), and 3) if the object category label is coarse, we go one level deeper into the label hierarchy in SUNCG, and find a suitable sub-category (such as washer, \etc). These criteria lead to 8 categories out of 80, including bed, chair, desk, dresser, fridge, sofa, table and washer.

\xhdr{Initial Location and Viewpoint.} We first define the \emph{visibility} of object by the ratio between visible and amodal masks. Then, we randomly sampling spawning locations and viewpoints for the agent by following: 1) The agent should be spawned close to the object, within distances between 3 to 6 meters; 2) The object visibility should be no less than $0.2$; 3) At most 6 instances are sampled for each object category in one house. Finally, we obtain 8940 instances in training set, 1113 in validation set and 2170 in test set. We also categorize spawning locations into ``hard'' instances if the object visibility is less than $0.5$; otherwise ``easy''. In \figref{fig:dataset}, each column shows an example ground-truth annotation (top) and the agent's initial location, viewpoint and the target's location (bottom). From left to right, they are easy, hard and partially out-of-view samples. In \figref{fig:obj_stats}, we further provide a summary of object occurrences in our dataset. It shows that our dataset is relatively balanced across different categories and difficulties.

\xhdr{Action Space.} We configure our agent with two sets of primitive actions: moving and turning. For moving, we allow agents to \emph{move forward, backward, left}, and \emph{right} without changing viewing angle. For turning, we allow agents to \emph{turn left} or \emph{right} for 2 degrees. This results in six actions in the action space. Note that we include \emph{move backward} in the action space considering that agent might need to backward to get rid of the occlusions.

\xhdr{Shortest Paths.} %
Since EVR aims to learn to move around to recognize occluded objects better, it is \emph{not} immediately clear what the ``ground-truth'' moving path is. This is different from other tasks, \eg point navigation, where the shortest path can serve as an ``oracle'' proxy. Nevertheless, as shortest-path navigation allows the agent to move closer to the target object and likely gain a better view, we still provide shortest-paths as part of our dataset, hoping it can provide both imitation supervision and a strong baseline. %

\section{Embodied Mask R-CNN}
\label{sec:embodiedmaskrcnn}
In this section, we propose a model called \emph{Embodied Mask R-CNN} to address the Embodied Visual Recognition. The proposed model consists of two modules, visual recognition module and action module, as outlined in Fig.~\ref{fig:framework}.

Before discussing the detailed designs, we first define the notations. The agent is spawned with initial location and gaze described in the previous section. Its initial observation of the environment is denoted by $I_0$, and the task specifies a target object with a bounding box $\bm{b}_0$ encompassing the visible region. Given the target object, the agent moves in the 3D environment following an action policy $\pi$. At each step $0$ to $T$, the agent takes action $a_t$ based on $\pi$ and observes an image $I_t$ from a view angle $\bm{v}_t$. The agent outputs its prediction of the object category, amodal bounding box and mask, denoted by $\bm{y}_t = \{c_t, \bm{b}_t, \bm{m}_t\}$ for the target object in the first frame. The goal is to recover the true object category, amodal bounding box, and amodal segmentation mask, $\bm{y}*=\{c^*, \bm{b}^{*}, \bm{m}^{*}\}$ at time step $0$.

\subsection{Visual Recognition}
\label{visual_perception}

The visual recognition module is responsible for predicting the object category, amodal bounding box, and amodal mask at each navigational time step. 

\xhdr{Mask R-CNN w/ Target Object.} Our visual recognition module has a similar goal as Mask R-CNN~\cite{he2017mask}, so we followed the architecture design. In our task, since the agent is already provided with the visible location of target object in the first frame, we remove the region proposal network from Mask R-CNN and directly use the location box to feed into the second stage. In our implementation, we use ResNet-50~\cite{he2016deep} pre-trianed on ImageNet as the backbone.


\xhdr{Temporal Mask R-CNN.} Given the sequential data along agent's trajectory, Temporal Mask R-CNN aims at aggregating temporal information from multiple frames to obtain more accurate predictions. Formally, the prediction of our temporal Mask R-CNN at time step $t$ is:
\begin{equation}
     \bm{y}_t = f(\bm{b}_0, I_0, I_1,...,I_t).
     \label{Eq:Temporal-mask-rcnn}
\end{equation}

The challenge is how to aggregate information from $\{I_0, I_1, \ldots, I_t\}$ together, especially when the 3D structure of the scene and the locations of the target object in the later frames are not unknown. To address this problem, we use feature-level fusion. 

Our perception model has three components: $\{f_\text{base}, f_\text{fuse}, f_\text{head}\}$.
For each frame $I_t$, we first use a convolutional neural network to extract a feature map $\bm{x}_t = f_\text{base}(I_t)$. Then, a feature aggregation function combines all the feature map up to $t$, resulting in a fused feature map $\hat{\bm{x}}_t = f_\text{fuse}(\bm{x}_0, \ldots, \bm{x}_t)$. For the feature aggregation model $f_\text{fuse}$, we use a single-layer Recurrent Convolution Network with Gated Recurrent Unit (GRU-RCN)~\cite{chung2014empirical,ballas2016delving} to fuse temporal features. These features are then sent to a Region-of-Interest (RoI)~\cite{girshick2015fast} head layer $f_\text{head}$ to make predictions for the first frame.
\begin{equation}
    \bm{y}_t = f_\text{head}(\bm{b}_0, \hat{\bm{x}}_t).
\end{equation}

\begin{figure}[t]
	\centering
	\includegraphics[width=1.0\linewidth]{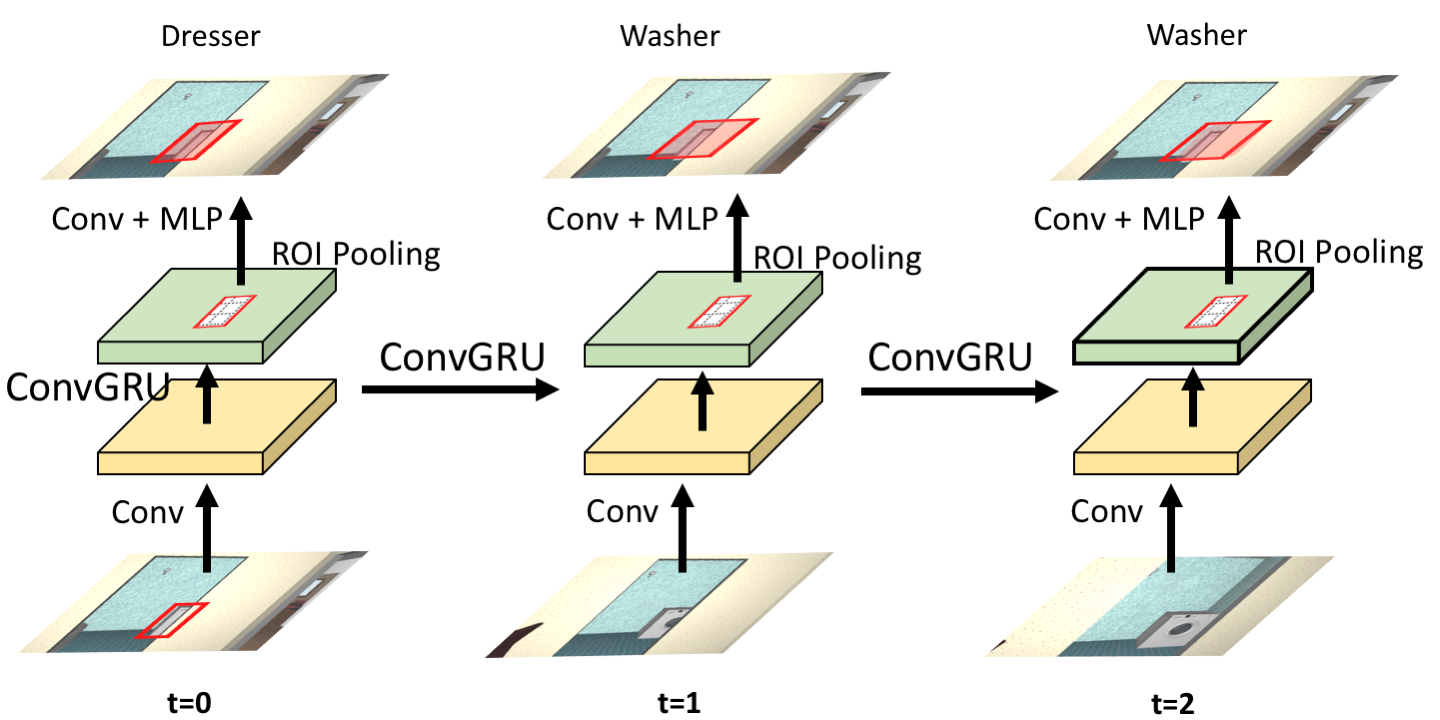}
	\vspace{-15pt}
    \caption{The visual recognition part of Embodied Mask R-CNN. The agent moves in the environment, acquires different views in each step (bottom row), and updates the prediction for the target object of the first frame (top row).}
	\label{fig:perception}
\end{figure}

To train the model, we use image sequences generated from the shortest-path trajectory. The overall loss for our visual recognition is defined as:
\begin{equation}
    L^p = \frac{1}{T} \sum_{t=1}^{T} \Big[L^p_c(c_t, c^*) + L^p_b(\bm{b}_t, \bm{b}^*) + L^p_m(\bm{m}_t, \bm{m}^*)\Big],
\end{equation}
where $L^p_{c}$ is the cross-entropy loss, $L^p_{b}$ is the smooth L1 regression loss, and $L^p_{m}$ is the binary cross-entropy loss~\cite{he2017mask}.

\subsection{Learning to Move}
\label{sec:rl_policy}

The goal of the policy network is to propose the next moves in order to acquire useful information for visual recognition. We disentangle it with the perception network, so that the learned policy will not over-fit to a specific perception model. We elaborate our design as follows.

\xhdr{Policy Network.} Similar to the perception network, the policy network receives a visible bounding box of target object $\bm{b}_0$ and the raw images as inputs, and outputs probabilities over the action space. We sample actions at step $t$ using:
\begin{equation}
    a_t \sim \pi(\bm{b}_0, I_0,I_1,...I_t).
\end{equation}

The policy network has three components $\{\pi_\text{imgEnc},\pi_\text{actEnc},\pi_\text{act}\}$. 
$\pi_\text{imgEnc}$ is an encoder for image features. The inputs $I_0$, $I_t$, as well as a mask $I^b$ representing the visible bounding box of the target object $\bm{b}_0$ in the initial view. We concatenate those inputs, resize them to $320{\times}384$, and pass them to $\pi_\text{imgEnc}$, which consists of four $\{5{\times}5~\text{Conv}, \text{BatchNorm}, \text{ReLU}, 2{\times}2~\text{MaxPool}\}$ blocks~\cite{embodiedqa}, producing an encoded image feature:
$\bm{z}^{\text{img}}_t = \pi_\text{imgEnc}\Big([I^b, I_0, I_t]\Big)$.

Besides image features, we also encode the last action in each step $t$. We use a multi-layer embedding network $\pi_\text{actEnc}$, which consists of an embedding layer, producing an encoded action feature $\bm{z}^\text{act}_t = \pi_\text{actEnc}(a_{t-1})$. We concatenate $\bm{z}^\text{act}_t$ and $\bm{z}^{\text{img}}_t$, and pass it to a single-layer GRU network $\pi_\text{act}$, whose output is sent to a linear layer with SoftMax to obtain a probability distribution over the action space:
\begin{equation}
    a_t \sim \pi_\text{act}\Big([\bm{z}^\text{img}_t, \bm{z}^\text{act}_t]\Big).
\end{equation}

\begin{figure}[t]
    \centering
	\includegraphics[width=1.0\linewidth]{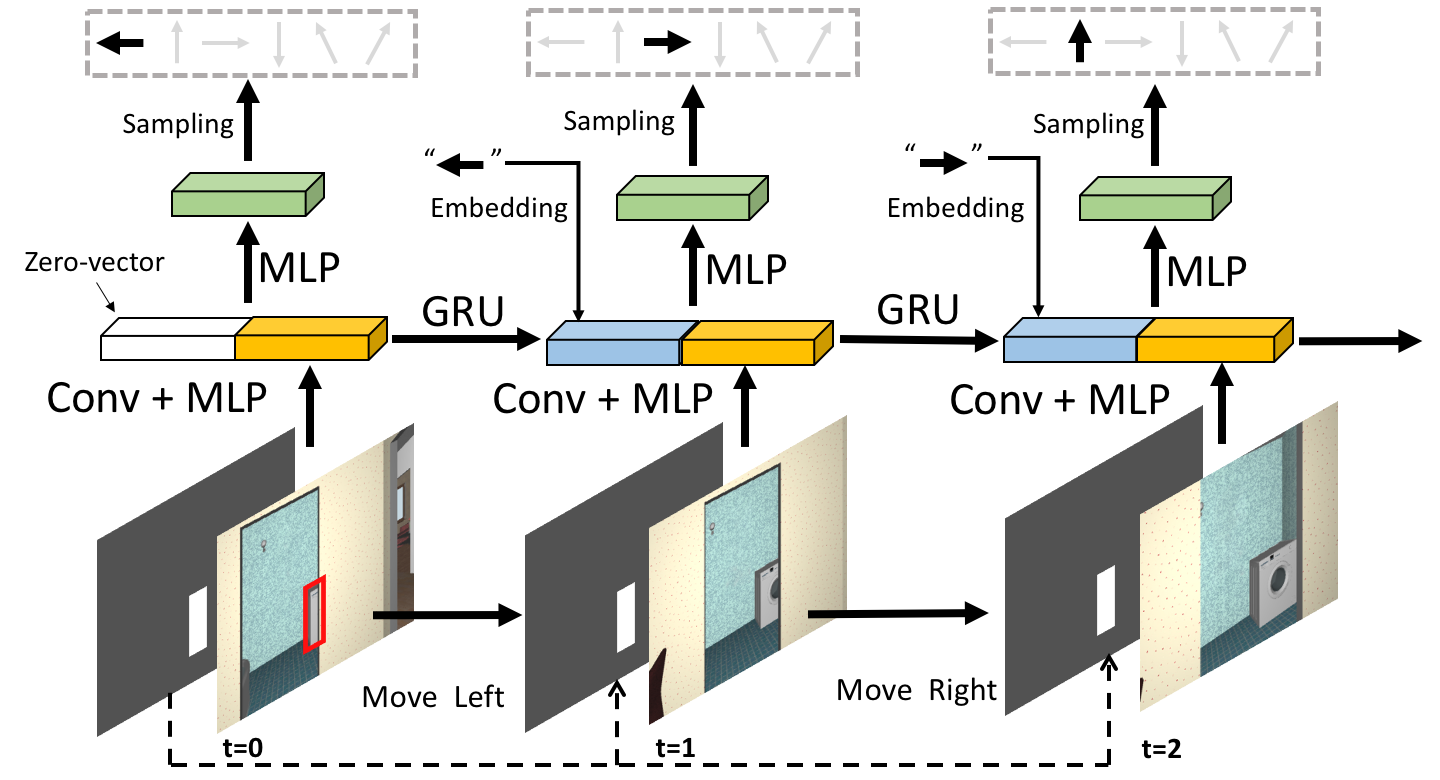}
    \caption{The action policy part of Embodied Mask R-CNN. At each step, the agent takes the current visual observation, last action and initial visible bounding box of target object as input, and predicts which action to take.} 
	\label{fig:network_policy}
\end{figure}

We learn $\{\pi_\text{imgEnc},\pi_\text{actEnc},\pi_\text{act}\}$ via reinforcement learning. We now describe how we design the reward.

\xhdr{Rewards.} Our goal is to find a good strategy for the agent to move to improve its visual recognition performance. We directly use the classification accuracy and Intersection-over-Union (IoU) to measure the advantages of candidate agent moves. Specifically, at each step $t$, we obtain the prediction of visual recognition $\bm{y}_t$, and then compute the classification accuracy $Acc^{c}_t$ (1 if correct, otherwise 0), and IoU between the amodal bounding box $IoU^{b}_t$ and mask $IoU^{m}_t$. Due to the different scales of these three rewards, we perform a weighted sum and then use reward shaping:
\begin{align}
    r_t &= \lambda_c {Acc}^c_t  + \lambda_b {IoU}^b_t + \lambda_b {IoU}^m_t, \\
    R_t &= r_t - r_{t-1},
\end{align}
where $\lambda_c{=}0.1$, $\lambda_b{=}10$ and $\lambda_m{=}20$. To learn the policy network $\pi$, we use policy gradient with REINFORCE \cite{sutton1998introduction}.

\subsection{Staged Training}
\label{sec:iterative_training}
We observe that joint training of the perception and policy networks from scratch struggles because the perception model cannot provide a correct reward to the policy network, and the policy network cannot take reasonable actions in turn. We thus resort to an staged training strategy. Namely, we first train the perception network with frames collected from the shortest path. Then, we plug in the pre-trained perception network to train the policy network with the perception part fixed. Finally, we retrain the perception network so that it can adapt to the learned action policy.
\section{Experiments}
\label{sec:experiments}

\subsection{Metrics and Baselines}
\vspace{-5pt} \xhdr{Metrics.} 
Recall that we evaluate the visual recognition performance on the first frame in the moving path. We report object classification accuracy (Clss-Acc), and the IoU scores for amodal box (ABox-IoU) and amodal mask (AMask-IoU). We additionally evaluate the performance of amodal segmentation \emph{only} on the occluded region of the target object (AMask-Occ-IoU).

\xhdr{Baselines.} 
We conduct extensive comparisons against a number of baselines. 
We use the format Training/Testing moving paths to characterize the baselines. 
\begin{compactitem}
    \item \textit{Passive/Passive (PP/PP)}: This is the conventional passive visual recognition setting, where the agent does not move during training and testing. The comparison to this baseline establishes the benefit of embodiment. 
    
    \item \textit{ShortestPath/Passive (SP/PP)}: The agent moves along the shortest path for training visual recognition, but the agent does not move during testing. We use this baseline to understand how much improvement is due to additional unlabeled data. 
    
    \item \textit{ShortestPath/Passive* (SP/PP*)}: Training is the same as above; In testing, the agent does not move, but we replicate the initial frame to create fake observations along the moving path to feed to the model. This baseline determines whether the improvement is due to the effectiveness of the recurrent network.

    \item \textit{ShortestPath/RandomPath (SP/RP)}: The agent moves randomly during test. This baseline establishes whether strategic move is required for embodied visual recognition. We report the performance by taking the average of scores of five random tests.
    
    \item \textit{ShortestPath/ShortestPath (SP/SP)}: The agent moves along the shortest path during both training and testing. 
    This is an ``oracle-like'' baseline, because in order to construct shortest-path, the agent need to know the entire 3D structure of the scene. However, 
    there is no guarantee that this is an optimal path 
    for recognition. 

\end{compactitem}
We compare these baselines with our two final models: \textit{ShortestPath/ActivePath (SP/AP)} and \textit{ActivePath/ActivePath (AP/AP)}. For \textit{ShortestPath/ActivePath}, we train the visual recognition model using frames in shortest path trajectories, and then train our own action policy. For \textit{ActivePath/ActivePath}, we further fine-tune our visual recognition model using rendered images generated from the learned action policy.

Noticeably, all the above models use the same temporal Mask R-CNN architecture for visual recognition. For single-frame prediction, the GRU module is also present. 
Moreover, all of those models are trained using the same amount of supervision and then evaluated on the same test set for fair comparison. For simplicity, we use the short name to represent each method in the figures.

\subsection{Implementation Details}

\begin{table*}[!t]
    \centering
    \begin{tabular}{c c c c c c c c c c c c c c}
    \toprule
    \multicolumn{2}{c}{Moving Path} & 
    \multicolumn{3}{c}{{Clss-Acc}} & \multicolumn{3}{c}{{ABox-IoU}} &  \multicolumn{3}{c}{{AMask-IoU}} & \multicolumn{3}{c}{{AMask-Occ-IoU}}\\
    \cmidrule(r){1-2}
    \cmidrule(r){3-5}
    \cmidrule(r){6-8}
    \cmidrule(r){9-11}
    \cmidrule(r){12-14}
    Train & Test & {all} & easy & hard & all & easy & hard & all & easy & hard & all & easy & hard \\
    \midrule	  
        \small{Passive} & \small{Passive} & 92.9 & 94.1 & 90.9 & 81.3 & 83.9 & 76.5 & 67.6 & 69.6 & 63.9 & 49.0 & 46.0 & 54.6\\
        \small{ShortestPath} & \small{Passive} & 92.8 & 94.3 & 89.9 & 81.2 & 83.8 & 76.4 & 67.4 & 69.6 & 63.4 & 48.6 & 45.8 & 54.1 \\
        \small{ShortestPath} & \small{Passive*}  & 93.0 & 94.3 & 90.7 & 80.9 & 83.1 & 76.8 & 66.7 & 68.4 & 63.6 & 48.4 & 44.9 & 54.9 \\
        \cmidrule(r){1-14}
        \small{ShortestPath} & \small{RandomPath}  & 93.1 & 94.1 & 91.1 & 81.6 & 83.9 & 77.1 & 67.8 & 69.7 & 64.3 & 49.0 & 45.8 & 55.2  \\       
        \small{ShortestPath} & \small{ShortestPath}  & 93.2 & 94.1 & 91.7 & 82.0 & 84.3 & 77.7 & 68.6 & 70.4 & 65.3 & \textbf{50.2} & \textbf{46.9} & 56.3 \\
        \small{ShortestPath} & \small{ActivePath}  & 93.3 & 93.9 & \textbf{92.2} & 82.0 & \textbf{84.4} & 77.6 & \textbf{68.8} & \textbf{70.5} & 65.5 & \textbf{50.2} & \textbf{46.9} & 56.4 \\
        \small{ActivePath} & \small{ActivePath}  & \textbf{93.7} & \textbf{94.6} & \textbf{92.2} & \textbf{82.2} & 84.3 & \textbf{78.2} & 68.7 & 70.3 & \textbf{65.6} & \textbf{50.2} & 46.8 & \textbf{56.7}  \\
    \bottomrule
    \end{tabular}
    \vspace{-5pt}
    \caption{Quantitative comparison of visual recognition using different models. ``Train'' denotes the source of moving path used to train the perception model; ``Test'' denotes the moving path in the testing stage. We report the performance at last (10-th) action step for embodied agents.}
    \label{table:results_all}\vspace{-2mm}
\end{table*}
\label{sec:experiments_implementation}

Here we provide the implementation details of our full system \textit{ActivePath/ActivePath}. There are three stages:

\xhdr{Stage 1: training visual recognition.} 
We implement our visual recognition model, Temporal Mask R-CNN, based on the PyTorch implementation of Mask R-CNN~\cite{massa2018mrcnn}. We use ResNet50~\cite{he2016deep} pre-trained from ImageNet~\cite{ImageNet} as the backbone and crop RoI features with a C4 head~\cite{ren2015faster}. The first three residual blocks in the backbone are fixed during training. We use stochastic gradient descent (SGD) with learning rate $0.0025$, batch size $8$, momentum $0.99$, and weight decay $0.0005$. 

\xhdr{Stage 2: training action policy.} We fix the visual recognition model, and train the action policy \textit{from scratch}. We used RMSProp~\cite{hinton2012neural} as the optimizer with initial learning rate $0.00004$, and set $\epsilon{=}0.00005$. In all our experiments, the agent moves 10 steps in total.

\xhdr{Stage 3: fine-tuning visual recognition.} Based on the learned action policy, we fine-tune the visual recognition model, so that it can adapt to the learned moving path. We use SGD, with learning rate $0.0005$.

\subsection{General Analysis on Experimental Results}

\label{subsec:experiemnts_results}

In ~\tableref{table:results_all}, we show the quantitative comparison of visual recognition performance for different models. We report the numbers on all examples from the test set (`all'), the easy examples (visibility $>0.5$), and hard examples (visibility $\le 0.5$). We have the following observations. 

\xhdr{Shortest path move does not help passive visual recognition.} As shown in \tableref{table:results_all}, both \textit{ShortestPath/Passive} and \textit{ShortestPath/Passive*} are slightly inferior to \textit{Passive/Passive}. Due to the movement, the visual appearance of additional images might change a lot compared with the first frame. As such, these extra inputs  does not appear to serve as effective data augmentation for training visual recognition in passive vision systems.

\xhdr{Embodiment helps visual recognition.} In~\tableref{table:results_all}, we can find that agents that move at test time (bottom four rows) consistently outperform agents that stay still (first three rows). Interestingly, \emph{even moving randomly} at test time (\textit{ShortestPath/RandomPath}), the agent still outperforms passive one. This provides evidence for this embodied paradigm helps visual recognition and the proposed Embodied Mask R-CNN model is effective for EVR.

\begin{figure}[t]
    \centering
    \begin{tabular}{cc}
   	\includegraphics[width=0.48\linewidth]{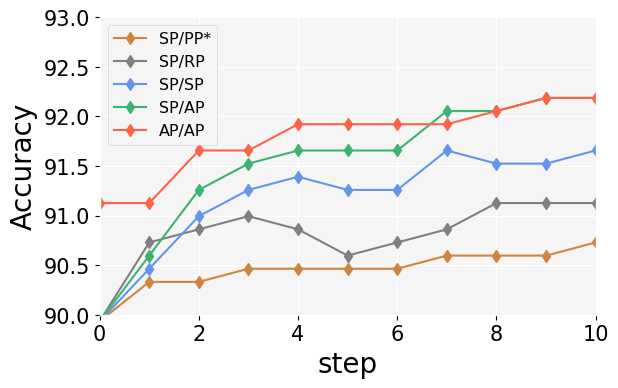}
   	 \includegraphics[width=0.48\linewidth]{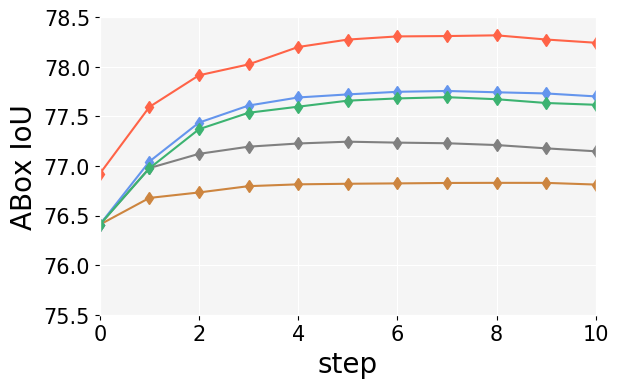} \\
   	\includegraphics[width=0.48\linewidth]{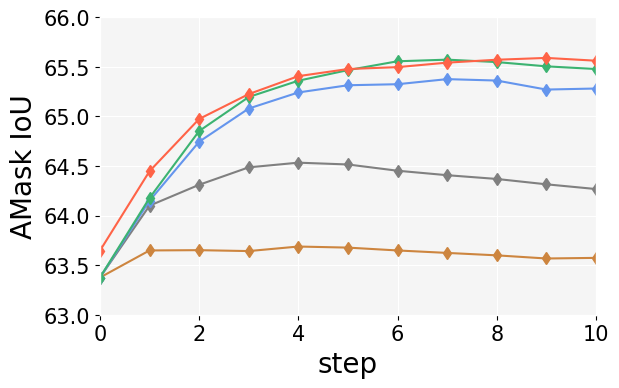}
   	 \includegraphics[width=0.48\linewidth]{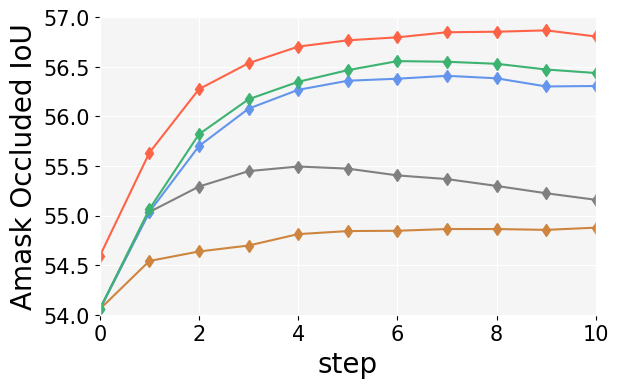} \\   	 
    \end{tabular}
    \vspace{-15pt}
    \caption{Performance of different models on hard samples over action step on four metrics.}
    \vspace{-10pt}
    \label{fig:step_perf}%
\end{figure}

\xhdr{Our model learns a better moving strategy.} In~\tableref{table:results_all}, we compare the models with embodiment (bottom four rows). The shortest path is derived to guide the agent move \emph{close} to the target object. It may not be the optimal moving strategy for EVR, since the task does not necessarily require the agent to get close to the target object. In contrast, our model learns a moving strategy to improve the agent's visual recognition ability. Though using the same visual recognition model, \textit{ShortestPath/ActivePath} finds a better moving strategy, and the performance is on par or slightly better than \textit{ShortestPath/ShortestPath}. After fine-tuning the visual recognition model using the learned path, \textit(ActivePath/ActivePath) achieves further improvement by adapting the visual recognition model to the learned paths.

\begin{figure}[t]
    \centering
   	\includegraphics[width=1.0\linewidth]{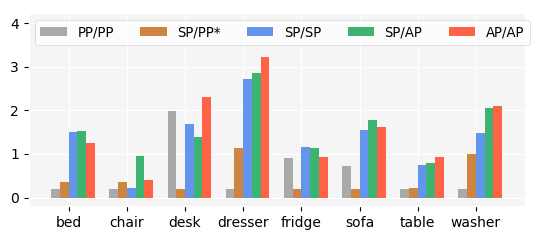}
    \vspace{-15pt}
    \caption{Relative comparison on different object categories for different methods. The numbers are obtained by taking the average of the first three metrics.}
    \vspace{-5pt}
    \label{fig:amask_barchart}
\end{figure}

\subsection{Analysis on Visual Recognition}

\vspace{-5pt} \xhdr{Objects with different occlusions}. In \tableref{table:results_all}, we observe that agents with embodiment in general achieve more improvement on ``hard'' samples compared with ``easy'' samples. For example, the object classification accuracy of \textit{ActivePath/ActivePath} is $0.5\%$ higher than \textit{Passive/Passive} for ``easy'' samples, while $1.3\%$ higher for ``hard'' samples. In general, objects with heavy occlusions are more difficult to recognize from single viewpoint, and embodiment helps because it can recover the occluded object portions.

\xhdr{Improvements over action step.} We show visual recognition performance along the action step in~\figref{fig:step_perf} on hard samples. In general, the performance improves as more steps are taken and more information aggregated, but eventually saturates. We suspect that the agent's location and viewpoint might change much after a number of steps, it becomes more challenging for it to aggregate information.

\xhdr{Performances on different object categories}. In Fig.~\ref{fig:amask_barchart}, we plot the relative improvements for different models on different object categories (we add a small constant value to each in the visualization for clarity). For comparison, we compute the average of the first three metrics for each category and all samples. The improvement is more significant on categories such as bed, dresser, sofa, table, and washer.

\subsection{Analysis on the Learned Policy}

\begingroup
\setlength{\tabcolsep}{1.6pt} %
   \begin{figure*}[!ht]
    \hspace{-1em}
     \begin{minipage}[t]{0.2\textwidth}
     \begin{tabular}{c}
     \subfigimg[width=1\linewidth]{Input Image}{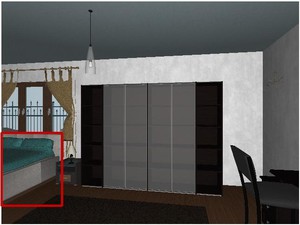} \\
     \subfigimg[width=1\linewidth]{Input Image}{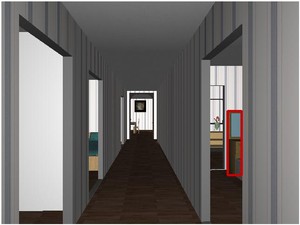}\\
     \subfigimg[width=1\linewidth]{Input Image}{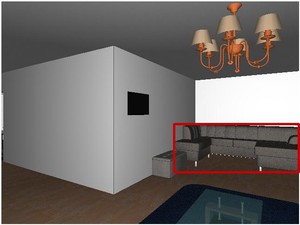}
	 \end{tabular}
     \end{minipage}
     \begin{minipage}[t]{0.8\textwidth}
       \begin{tabular}{|cccccc|cc}
        \subfigimg[width=0.12\linewidth]{Forward}{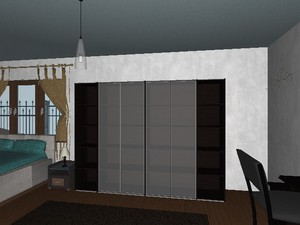}&
        \subfigimg[width=0.12\linewidth]{MoveLeft}{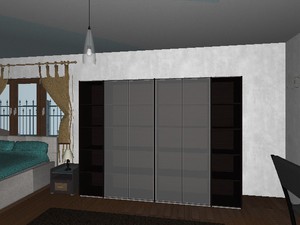}&
        \subfigimg[width=0.12\linewidth]{MoveLeft}{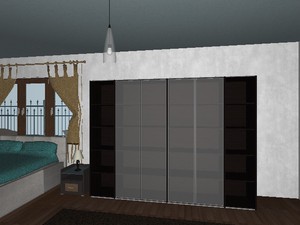}&
        \subfigimg[width=0.12\linewidth]{MoveLeft}{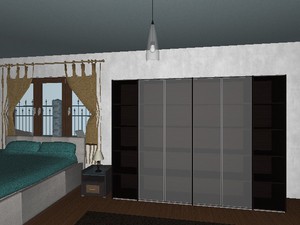}&
        \subfigimg[width=0.12\linewidth]{RotateLeft}{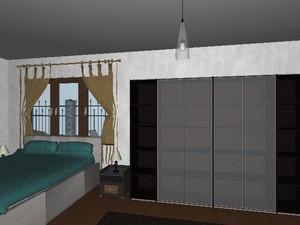}&
        \subfigimg[width=0.12\linewidth]{RotateLeft}{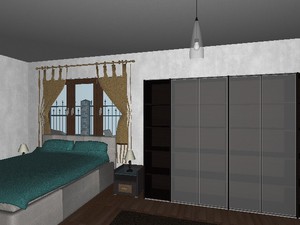}&
        \subfigimg[width=0.12\linewidth]{Output-SP}{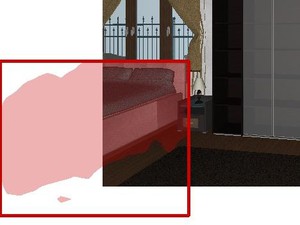}&
        \subfigimg[width=0.12\linewidth]{GroundTruth}{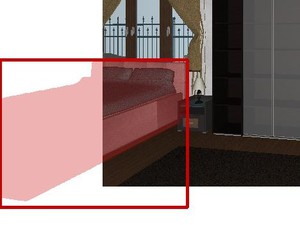}\\
        \subfigimg[width=0.12\linewidth]{RotateLeft}{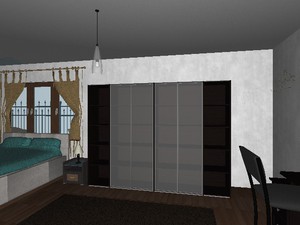}&
        \subfigimg[width=0.12\linewidth]{RotateLeft}{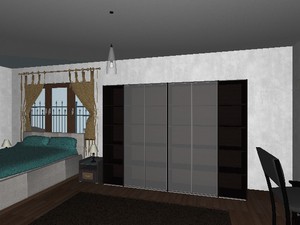}&
        \subfigimg[width=0.12\linewidth]{RotateLeft}{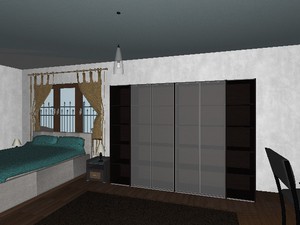}&
        \subfigimg[width=0.12\linewidth]{Backward}{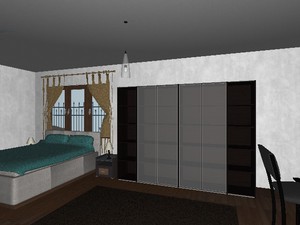}&
        \subfigimg[width=0.12\linewidth]{Backward}{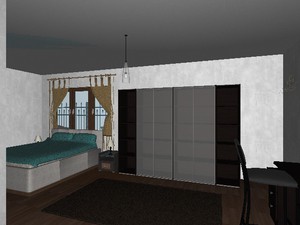}&
        \subfigimg[width=0.12\linewidth]{Backward}{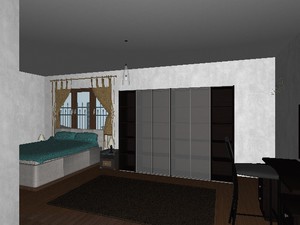}&
        \subfigimg[width=0.12\linewidth]{Output-AP}{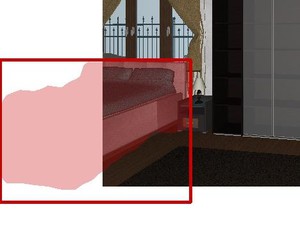}&
        \subfigimg[width=0.12\linewidth]{GroundTruth}{figs/experiment/visualizations/sample4/4_gt_crop.jpg}\\

        \subfigimg[width=0.12\linewidth]{RotateRight}{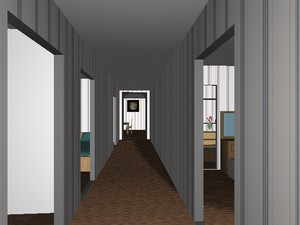}&
        \subfigimg[width=0.12\linewidth]{RotateRight}{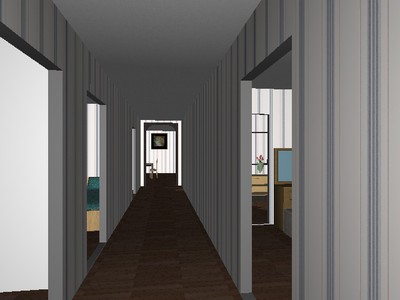}&
        \subfigimg[width=0.12\linewidth]{RotateRight}{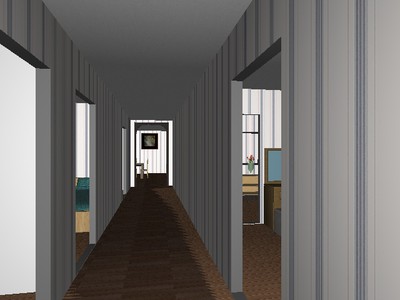}&
        \subfigimg[width=0.12\linewidth]{Forward}{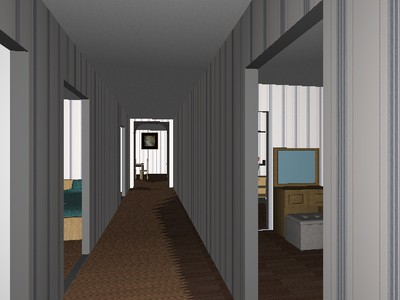}&
        \subfigimg[width=0.12\linewidth]{RotateRight}{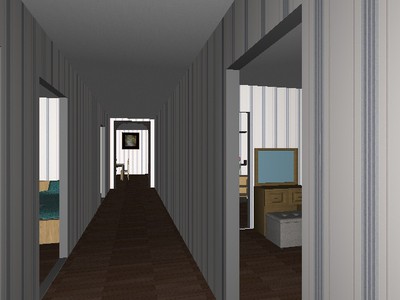}&
        \subfigimg[width=0.12\linewidth]{RotateRight}{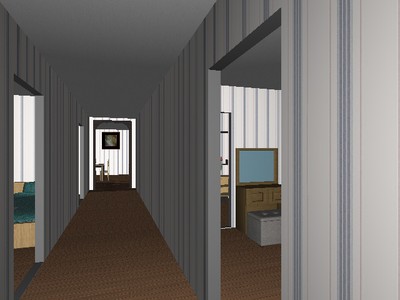}&
        \subfigimg[width=0.12\linewidth]{Output-SP}{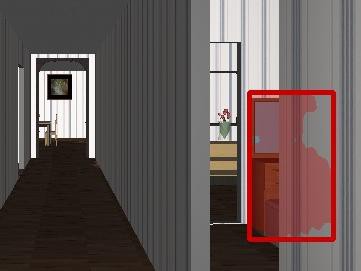}&
        \subfigimg[width=0.12\linewidth]{GroundTruth}{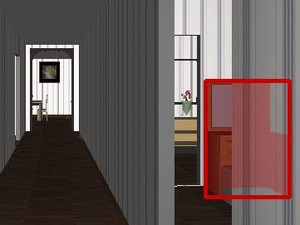}\\
        \subfigimg[width=0.12\linewidth]{RotateRight}{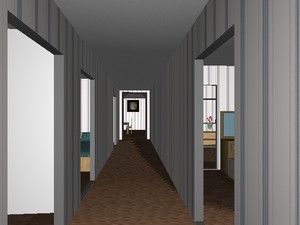}&
        \subfigimg[width=0.12\linewidth]{RotateRight}{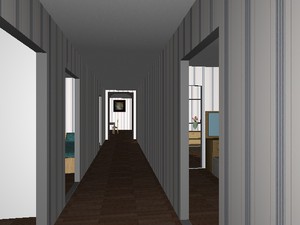}&
        \subfigimg[width=0.12\linewidth]{RotateRight}{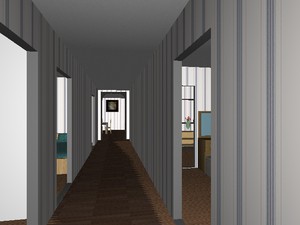}&
        \subfigimg[width=0.12\linewidth]{MoveLeft}{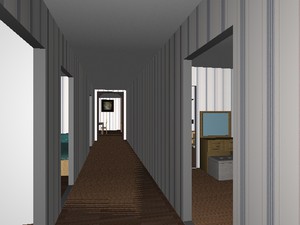}&
        \subfigimg[width=0.12\linewidth]{MoveLeft}{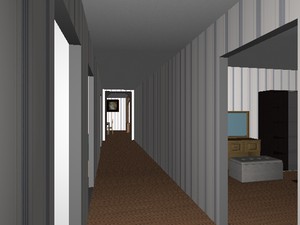}&
        \subfigimg[width=0.12\linewidth]{MoveLeft}{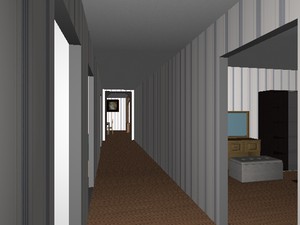}&
        \subfigimg[width=0.12\linewidth]{Output-AP}{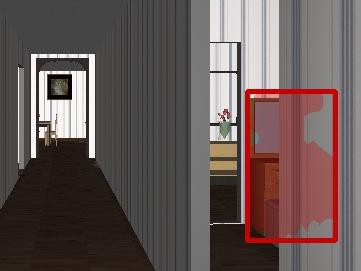}&
        \subfigimg[width=0.12\linewidth]{GroundTruth}{figs/experiment/visualizations/sample6/6_gt_crop.jpg}\\
        
        \subfigimg[width=0.12\linewidth]{RotateLeft}{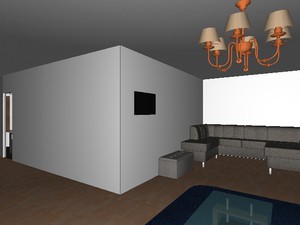}&
        \subfigimg[width=0.12\linewidth]{Forward}{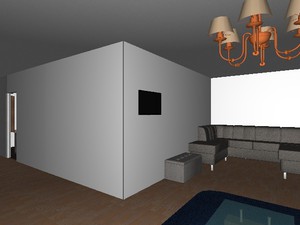}&
        \subfigimg[width=0.12\linewidth]{RotateLeft}{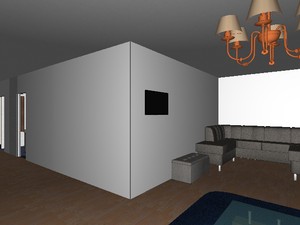}&
        \subfigimg[width=0.12\linewidth]{RotateRight}{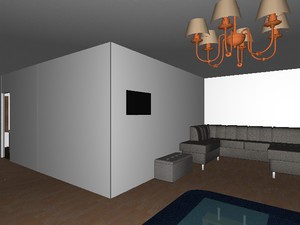}&
        \subfigimg[width=0.12\linewidth]{Forward}{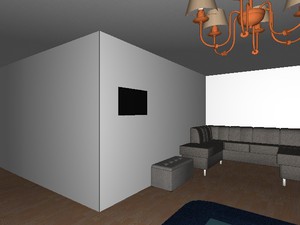}&
        \subfigimg[width=0.12\linewidth]{MoveRight}{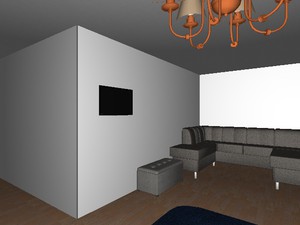}&
        \subfigimg[width=0.12\linewidth]{Output-SP}{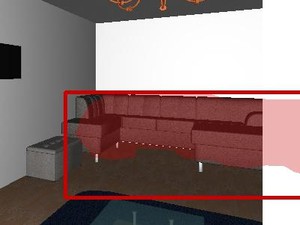}&
        \subfigimg[width=0.12\linewidth]{GroundTruth}{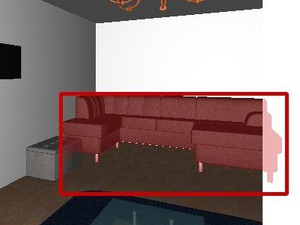}\\
        \subfigimg[width=0.12\linewidth]{RotateRight}{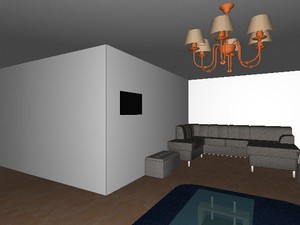}&
        \subfigimg[width=0.12\linewidth]{RotateRight}{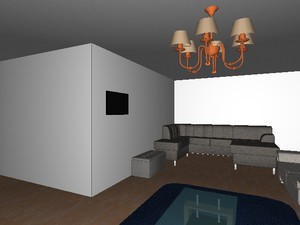}&
        \subfigimg[width=0.12\linewidth]{RotateRight}{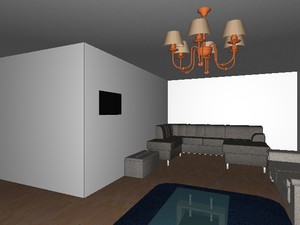}&
        \subfigimg[width=0.12\linewidth]{MoveRight}{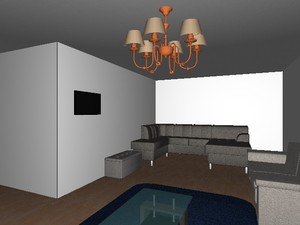}&
        \subfigimg[width=0.12\linewidth]{MoveLeft}{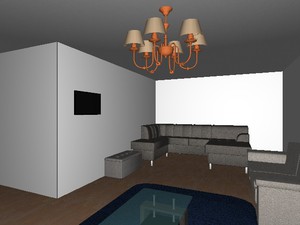}&
        \subfigimg[width=0.12\linewidth]{RotateLeft}{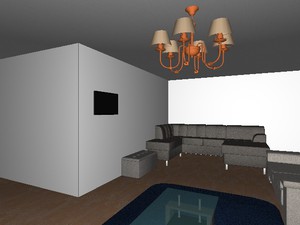}&
        \subfigimg[width=0.12\linewidth]{Output-AP}{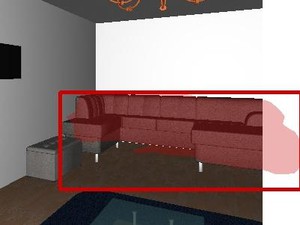}&
        \subfigimg[width=0.12\linewidth]{GroundTruth}{figs/experiment/visualizations/sample24/24_gt_crop.jpg}\\

       \end{tabular}
     \end{minipage}
     \vspace{-5pt}
     \caption{For each image, we visualize the shortest-path trajectory (top) and the learned active perception trajectory (bottom) at step 1, 3, 4, 6, 8, 10. Different from agents using the shortest-path move, our agents actively perceive the target object, and achieve better visual recognition performance.}
     \label{fig:visualization}\vspace{-1mm}
   \end{figure*}
\endgroup

\begin{figure}[t]
    \centering
    \begin{tabular}{c}
   	\includegraphics[width=0.98\linewidth, trim={1cm 0.8cm 0 0.8cm},clip]{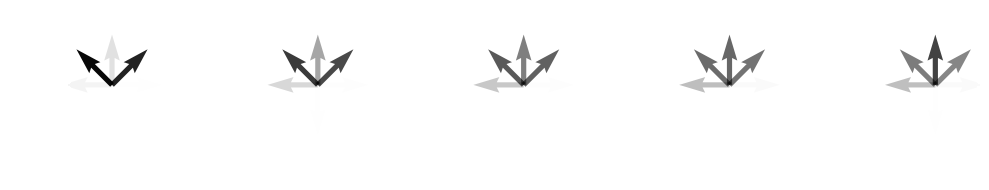}\\
   	\includegraphics[width=0.98\linewidth, trim={1cm 1cm 0 1cm},clip]{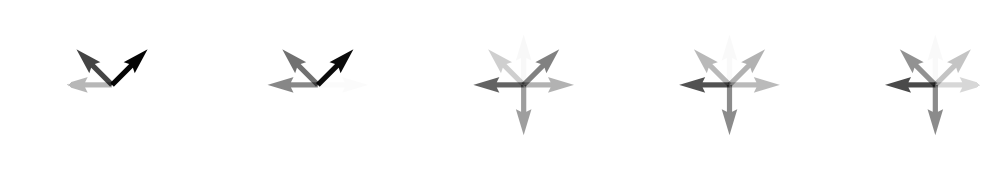}
    \end{tabular}
    \vspace{-10pt}
    \caption{Distribution of actions at step {1, 3, 5, 7, 10} on test set. $\uparrow$: Forward, $\downarrow$: Backward, 	$\leftarrow$: Move left, $\rightarrow$: Move right, $\nwarrow$: Rotate left, $\nearrow$: Rotate right. Top row: shortest-path movement. Bottom row: our learned policy. Darker color denotes more frequent actions.}
    \label{fig:action_dist}
\end{figure}

We visualize some example moving paths executed by learned policy 
(\textit{ActivePath/ActivePath}) in~\figref{fig:visualization}. Using the learned moving paths, agents can predict better amodal masks compared with shortest path, and their moving patterns are also different.

\xhdr{Comparing moving strategies}. \figref{fig:action_dist} shows the distribution of actions at steps 1, 3, 5, 7 and 10 for the shortest path and our learned path. We can observe different moving strategies are learned from our model compared with shortest path even though the visual recognition model is shared by two models. Specifically, our agent rarely moves forward. Instead, it learns to occasionally move backward. This comparison indicates the shortest path may not be the optimal path for EVR.

\begin{figure}[t]
    \centering
   	\includegraphics[width=1\linewidth, trim={0 0.2cm 0 0.2cm}, clip]{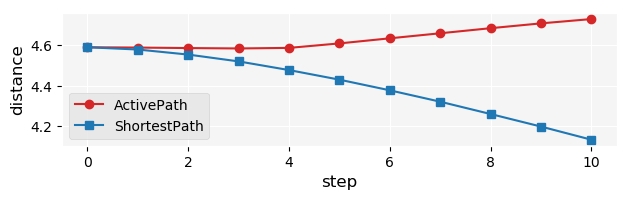}
   	\vspace{-15pt}
    \caption{Distance to target objects at each step averaged on test set for shortest path and our learned path.}
    \label{fig:step_vis_dist}
    \vspace{-10pt}
\end{figure}

\xhdr{Distance to the target object}. We further investigate the moving strategy in terms of the distance to the target object. As shown in Fig.~\ref{fig:step_vis_dist}, under the shortest path, the agent gets closer to the object. However, our learned moves keep the distance nearly constant to the target object. Under this moving strategy, the viewed-size of the target object at each step does not change too drastically. This can be beneficial in cases where the agent is spawned close to the target, and moving backward can reveal more content of the object.
\section{Conclusion}
In this work, we introduced a new task called \emph{Embodied Visual Recognition}\,---\,an agent is spawned in a 3D environment, and is free to move in order to perform object classification, amodal localization and segmentation of a target occluded object. As a first step toward solving this task, we proposed an \emph{Embodied Mask R-CNN} model that learned to move strategically to improve the visual recognition performance. Through comparisons with various baselines, we demonstrated the importance of embodiment for visual recognition. We also show that our agents developed strategical movements that were different from shortest path, to recover the semantics and shape of occluded objects.

\xhdr{Acknowledgments.} We thank Manolis Savva, Marcus Rohrbach, and Lixing Liu for the helpful discussions about task formulation, and Dipendra Misra for helpful RL training tips. This work was supported in part by NSF (CAREER IIS-1253549), AFRL, DARPA, Siemens, Samsung, Google, Amazon, ONR YIPs, ONR Grants N00014-16-1-\{2713,2793\}, 
the IU Office of the Vice Provost for Research, and the College of Arts and Sciences, the School of Informatics, Computing, and
Engineering through the Emerging Areas of Research Project ``Learning: Brains, Machines, and Children.''. The views and conclusions contained herein are those of the authors and should not be interpreted as necessarily representing the official policies or endorsements, either expressed or implied, of the U.S. Government, or any sponsor.

{\small
\bibliographystyle{ieee}
\bibliography{evr}

\begin{thebibliography}{10}\itemsep=-1pt

\bibitem{aloimonos1988active}
J.~Aloimonos, I.~Weiss, and A.~Bandyopadhyay.
\newblock Active vision.
\newblock {\em International Journal of Computer Vision (IJCV)}, 1988.

\bibitem{ammirato2017dataset}
P.~Ammirato, P.~Poirson, E.~Park, J.~Ko{\v{s}}eck{\'a}, and A.~C. Berg.
\newblock A dataset for developing and benchmarking active vision.
\newblock In {\em 2017 IEEE International Conference on Robotics and Automation
  (ICRA)}, 2017.

\bibitem{mattersim}
P.~Anderson, Q.~Wu, D.~Teney, J.~Bruce, M.~Johnson, N.~S{\"u}nderhauf, I.~Reid,
  S.~Gould, and A.~van~den Hengel.
\newblock {Vision-and-Language Navigation}: Interpreting visually-grounded
  navigation instructions in real environments.
\newblock In {\em Proceedings of the IEEE Conference on Computer Vision and
  Pattern Recognition (CVPR)}, 2018.

\bibitem{2017armeni2d3d}
I.~{Armeni}, A.~{Sax}, A.~R. {Zamir}, and S.~{Savarese}.
\newblock Joint {2D}-{3D}-semantic data for indoor scene understanding.
\newblock {\em arXiv preprint arXiv:1702.01105}, 2017.

\bibitem{Aubry14}
M.~Aubry, D.~Maturana, A.~Efros, B.~Russell, and J.~Sivic.
\newblock Seeing 3d chairs: exemplar part-based 2d-3d alignment using a large
  dataset of cad models.
\newblock In {\em Proceedings of the IEEE Conference on Computer Vision and
  Pattern Recognition (CVPR)}, 2014.

\bibitem{baillargeon1985object}
R.~Baillargeon, E.~S. Spelke, and S.~Wasserman.
\newblock Object permanence in five-month-old infants.
\newblock {\em Cognition}, 20(3):191--208, 1985.

\bibitem{bajcsy1988active}
R.~Bajcsy.
\newblock Active perception.
\newblock {\em Proceedings of the IEEE}, 1988.

\bibitem{ballas2016delving}
N.~Ballas, L.~Yao, C.~Pal, and A.~Courville.
\newblock Delving deeper into convolutional networks for learning video
  representations.
\newblock {\em International Conference on Lefvarning Representations (ICLR)},
  2016.

\bibitem{bambach2018toddler}
S.~Bambach, D.~J. Crandall, L.~B. Smith, and C.~Yu.
\newblock Toddler-inspired visual object learning.
\newblock In {\em Advances in Neural Information Processing Systems (NIPS)},
  2018.

\bibitem{bansal2016marr}
A.~Bansal, B.~Russell, and A.~Gupta.
\newblock Marr revisited: 2d-3d alignment via surface normal prediction.
\newblock In {\em Proceedings of the IEEE Conference on Computer Vision and
  Pattern Recognition (CVPR)}, 2016.

\bibitem{bao2018cnn}
L.~Bao, B.~Wu, and W.~Liu.
\newblock Cnn in mrf: Video object segmentation via inference in a cnn-based
  higher-order spatio-temporal mrf.
\newblock In {\em Proceedings of the IEEE Conference on Computer Vision and
  Pattern Recognition (CVPR)}, 2018.

\bibitem{brodeur2017home}
S.~Brodeur, E.~Perez, A.~Anand, F.~Golemo, L.~Celotti, F.~Strub, J.~Rouat,
  H.~Larochelle, and A.~Courville.
\newblock {HoME}: A household multimodal environment.
\newblock {\em arXiv preprint arXiv:1711.11017}, 2017.

\bibitem{caicedo2015active}
J.~C. Caicedo and S.~Lazebnik.
\newblock Active object localization with deep reinforcement learning.
\newblock In {\em Proceedings of the IEEE International Conference on Computer
  Vision}, 2015.

\bibitem{Cheng2018visual}
R.~Cheng, A.~Agarwal, and K.~Fragkiadaki.
\newblock Reinforcement learning of active vision for manipulating objects
  under occlusions.
\newblock In {\em Conference on Robot Learning (CoRL)}, 2018.

\bibitem{chung2014empirical}
J.~Chung, C.~Gulcehre, K.~Cho, and Y.~Bengio.
\newblock Empirical evaluation of gated recurrent neural networks on sequence
  modeling.
\newblock {\em arXiv preprint arXiv:1412.3555}, 2014.

\bibitem{embodiedqa}
A.~Das, S.~Datta, G.~Gkioxari, S.~Lee, D.~Parikh, and D.~Batra.
\newblock {E}mbodied {Q}uestion {A}nswering.
\newblock In {\em Proceedings of the IEEE Conference on Computer Vision and
  Pattern Recognition (CVPR)}, 2018.

\bibitem{eqa_modular}
A.~Das, G.~Gkioxari, S.~Lee, D.~Parikh, and D.~Batra.
\newblock {N}eural {M}odular {C}ontrol for {E}mbodied {Q}uestion {A}nswering.
\newblock In {\em Conference on Robot Learning (CoRL)}, 2018.

\bibitem{denzler2002information}
J.~Denzler and C.~M. Brown.
\newblock Information theoretic sensor data selection for active object
  recognition and state estimation.
\newblock {\em IEEE Transactions on Pattern Analysis and Machine Intelligence
  (TPAMI)}, 2002.

\bibitem{stephen1999vision}
S.~E.~Palmer.
\newblock {\em Vision Science: Photons to Phenomenology}.
\newblock The MIT Press, 1999.

\bibitem{ehsani2018segan}
K.~Ehsani, R.~Mottaghi, and A.~Farhadi.
\newblock Segan: Segmenting and generating the invisible.
\newblock In {\em Proceedings of the IEEE Conference on Computer Vision and
  Pattern Recognition (CVPR)}, 2018.

\bibitem{follmann2018learning}
P.~Follmann, R.~K{\"o}nig, P.~H{\"a}rtinger, and M.~Klostermann.
\newblock Learning to see the invisible: End-to-end trainable amodal instance
  segmentation.
\newblock {\em arXiv preprint arXiv:1804.08864}, 2018.

\bibitem{geirhos2018imagenet}
R.~Geirhos, P.~Rubisch, C.~Michaelis, M.~Bethge, F.~A. Wichmann, and
  W.~Brendel.
\newblock Imagenet-trained cnns are biased towards texture; increasing shape
  bias improves accuracy and robustness.
\newblock {\em arXiv preprint arXiv:1811.12231}, 2018.

\bibitem{girshick2015fast}
R.~Girshick.
\newblock Fast {R-CNN}.
\newblock In {\em Proc. of the IEEE International Conference on Computer Vision
  (ICCV)}, 2015.

\bibitem{girshick2014rich}
R.~Girshick, J.~Donahue, T.~Darrell, and J.~Malik.
\newblock Rich feature hierarchies for accurate object detection and semantic
  segmentation.
\newblock In {\em Proceedings of the IEEE Conference on Computer Vision and
  Pattern Recognition (CVPR)}, 2014.

\bibitem{gonzalez2015active}
A.~Gonzalez-Garcia, A.~Vezhnevets, and V.~Ferrari.
\newblock An active search strategy for efficient object class detection.
\newblock In {\em Proceedings of the IEEE Conference on Computer Vision and
  Pattern Recognition}, 2015.

\bibitem{gordon2018iqa}
D.~Gordon, A.~Kembhavi, M.~Rastegari, J.~Redmon, D.~Fox, and A.~Farhadi.
\newblock Iqa: Visual question answering in interactive environments.
\newblock In {\em Proceedings of the IEEE Conference on Computer Vision and
  Pattern Recognition}, 2018.

\bibitem{gupta2015aligning}
S.~Gupta, P.~Arbel{\'a}ez, R.~Girshick, and J.~Malik.
\newblock Aligning 3d models to rgb-d images of cluttered scenes.
\newblock In {\em Proceedings of the IEEE Conference on Computer Vision and
  Pattern Recognition (CVPR)}, 2015.

\bibitem{han2019active}
X.~Han, H.~Liu, F.~Sun, and X.~Zhang.
\newblock Active object detection with multi-step action prediction using deep
  q-network.
\newblock {\em IEEE Transactions on Industrial Informatics}, 2019.

\bibitem{he2017mask}
K.~He, G.~Gkioxari, P.~Doll{\'a}r, and R.~Girshick.
\newblock Mask {R-CNN}.
\newblock In {\em Proc. of the IEEE International Conference on Computer Vision
  (ICCV)}, 2017.

\bibitem{he2016deep}
K.~He, X.~Zhang, S.~Ren, and J.~Sun.
\newblock Deep residual learning for image recognition.
\newblock In {\em Proceedings of the IEEE Conference on Computer Vision and
  Pattern Recognition (CVPR)}, 2016.

\bibitem{hinton2012neural}
G.~Hinton, N.~Srivastava, and K.~Swersky.
\newblock Neural networks for machine learning lecture 6a overview of
  mini-batch gradient descent.
\newblock 2012.

\bibitem{hu2018senet}
J.~Hu, L.~Shen, and G.~Sun.
\newblock Squeeze-and-excitation networks.
\newblock {\em Proceedings of the IEEE Conference on Computer Vision and
  Pattern Recognition (CVPR)}, 2018.

\bibitem{izadinia2017im2cad}
H.~Izadinia, Q.~Shan, and S.~M. Seitz.
\newblock Im2cad.
\newblock In {\em Proceedings of the IEEE Conference on Computer Vision and
  Pattern Recognition (CVPR)}, 2017.

\bibitem{jayaraman2015learning}
D.~Jayaraman and K.~Grauman.
\newblock Learning image representations tied to ego-motion.
\newblock In {\em Proc. of the IEEE International Conference on Computer Vision
  (ICCV)}, 2015.

\bibitem{jayaraman2018end}
D.~Jayaraman and K.~Grauman.
\newblock End-to-end policy learning for active visual categorization.
\newblock {\em IEEE Transactions on Pattern Analysis and Machine Intelligence
  (TPAMI)}, 2018.

\bibitem{johns2016pairwise}
E.~Johns, S.~Leutenegger, and A.~J. Davison.
\newblock Pairwise decomposition of image sequences for active multi-view
  recognition.
\newblock In {\em Proceedings of the IEEE Conference on Computer Vision and
  Pattern Recognition (CVPR)}, 2016.

\bibitem{kanizsa1979Organization}
G.~Kanizsa.
\newblock {\em Organization in vision: Essays on Gestalt perception}.
\newblock Praeger, 1979.

\bibitem{kar2015amodal}
A.~Kar, S.~Tulsiani, J.~Carreira, and J.~Malik.
\newblock Amodal completion and size constancy in natural scenes.
\newblock In {\em Proceedings of the IEEE Conference on Computer Vision and
  Pattern Recognition (CVPR)}, 2015.

\bibitem{ai2thor}
E.~Kolve, R.~Mottaghi, D.~Gordon, Y.~Zhu, A.~Gupta, and A.~Farhadi.
\newblock {AI2-THOR: An Interactive 3D Environment for Visual AI}.
\newblock {\em arXiv preprint arXiv:1712.05474}, 2017.

\bibitem{kragic2005vision}
D.~Kragic, M.~Bj{\"o}rkman, H.~I. Christensen, and J.-O. Eklundh.
\newblock Vision for robotic object manipulation in domestic settings.
\newblock {\em Robotics and autonomous Systems}, 2005.

\bibitem{krizhevsky2012imagenet}
A.~Krizhevsky, I.~Sutskever, and G.~E. Hinton.
\newblock Imagenet classification with deep convolutional neural networks.
\newblock In {\em Advances in Neural Information Processing Systems (NIPS)},
  2012.

\bibitem{li2016amodal}
K.~Li and J.~Malik.
\newblock Amodal instance segmentation.
\newblock In {\em Proceedings of the European Conference on Computer Vision
  (ECCV)}, 2016.

\bibitem{lin2014microsoft}
T.-Y. Lin, M.~Maire, S.~Belongie, J.~Hays, P.~Perona, D.~Ramanan,
  P.~Doll{\'a}r, and C.~L. Zitnick.
\newblock Microsoft {COCO}: Common objects in context.
\newblock In {\em Proceedings of the European Conference on Computer Vision
  (ECCV)}, 2014.

\bibitem{long2015fully}
J.~Long, E.~Shelhamer, and T.~Darrell.
\newblock Fully convolutional networks for semantic segmentation.
\newblock In {\em Proceedings of the IEEE Conference on Computer Vision and
  Pattern Recognition (CVPR)}, 2015.

\bibitem{malmir2015deep}
M.~Malmir, K.~Sikka, D.~Forster, J.~Movellan, and G.~W. Cottrell.
\newblock Deep q-learning for active recognition of germs: Baseline performance
  on a standardized dataset for active learning.
\newblock {\em Proceedings of the British Machine Vision Conference (BMVC)},
  2015.

\bibitem{massa2018mrcnn}
F.~Massa and R.~Girshick.
\newblock {maskrnn-benchmark: Fast, modular reference implementation of
  Instance Segmentation and Object Detection algorithms in PyTorch}.
\newblock \url{https://github.com/facebookresearch/maskrcnn-benchmark}, 2018.

\bibitem{mathe2016reinforcement}
S.~Mathe, A.~Pirinen, and C.~Sminchisescu.
\newblock Reinforcement learning for visual object detection.
\newblock In {\em Proceedings of the IEEE Conference on Computer Vision and
  Pattern Recognition}, 2016.

\bibitem{novotny2017learning}
D.~Novotny, D.~Larlus, and A.~Vedaldi.
\newblock Learning {3D} object categories by looking around them.
\newblock In {\em Proc. of the IEEE International Conference on Computer Vision
  (ICCV)}, 2017.

\bibitem{pathak2018learning}
D.~Pathak, Y.~Shentu, D.~Chen, P.~Agrawal, T.~Darrell, S.~Levine, and J.~Malik.
\newblock Learning instance segmentation by interaction.
\newblock In {\em Proceedings of the IEEE Conference on Computer Vision and
  Pattern Recognition Workshops}, 2018.

\bibitem{redmon2016you}
J.~Redmon, S.~Divvala, R.~Girshick, and A.~Farhadi.
\newblock You only look once: Unified, real-time object detection.
\newblock In {\em Proceedings of the IEEE Conference on Computer Vision and
  Pattern Recognition (CVPR)}, 2016.

\bibitem{yolov3}
J.~Redmon and A.~Farhadi.
\newblock {YOLOv3}: An incremental improvement.
\newblock {\em arXiv preprint arXiv:1804.02767}, 2018.

\bibitem{ren2015faster}
S.~Ren, K.~He, R.~Girshick, and J.~Sun.
\newblock Faster r-cnn: Towards real-time object detection with region proposal
  networks.
\newblock In {\em Advances in Neural Information Processing Systems (NIPS)},
  2015.

\bibitem{ImageNet}
O.~Russakovsky, J.~Deng, H.~Su, J.~Krause, S.~Satheesh, S.~Ma, Z.~Huang,
  A.~Karpathy, A.~Khosla, M.~Bernstein, A.~C. Berg, and L.~Fei-Fei.
\newblock {ImageNet Large Scale Visual Recognition Challenge}.
\newblock {\em International Journal of Computer Vision (IJCV)}, 2015.

\bibitem{savva2017minos}
M.~Savva, A.~X. Chang, A.~Dosovitskiy, T.~Funkhouser, and V.~Koltun.
\newblock {MINOS}: Multimodal indoor simulator for navigation in complex
  environments.
\newblock {\em arXiv preprint arXiv:1712.03931}, 2017.

\bibitem{song2016ssc}
S.~Song, F.~Yu, A.~Zeng, A.~X. Chang, M.~Savva, and T.~Funkhouser.
\newblock Semantic scene completion from a single depth image.
\newblock {\em Proceedings of the IEEE Conference on Computer Vision and
  Pattern Recognition (CVPR)}, 2017.

\bibitem{su2015multi}
H.~Su, S.~Maji, E.~Kalogerakis, and E.~Learned-Miller.
\newblock Multi-view convolutional neural networks for 3d shape recognition.
\newblock In {\em Proc. of the IEEE International Conference on Computer Vision
  (ICCV)}, 2015.

\bibitem{sun2018pix3d}
X.~Sun, J.~Wu, X.~Zhang, Z.~Zhang, C.~Zhang, T.~Xue, J.~B. Tenenbaum, and W.~T.
  Freeman.
\newblock Pix3d: Dataset and methods for single-image 3d shape modeling.
\newblock In {\em The IEEE Conference on Computer Vision and Pattern
  Recognition (CVPR)}, 2018.

\bibitem{sutton1998introduction}
R.~S. Sutton, A.~G. Barto, et~al.
\newblock {\em Introduction to reinforcement learning}.
\newblock MIT press Cambridge, 1998.

\bibitem{szegedy2015going}
C.~Szegedy, W.~Liu, Y.~Jia, P.~Sermanet, S.~Reed, D.~Anguelov, D.~Erhan,
  V.~Vanhoucke, and A.~Rabinovich.
\newblock Going deeper with convolutions.
\newblock In {\em Proceedings of the IEEE Conference on Computer Vision and
  Pattern Recognition (CVPR)}, 2015.

\bibitem{wagemans2012century}
J.~Wagemans, J.~H. Elder, M.~Kubovy, S.~E. Palmer, M.~A. Peterson, M.~Singh,
  and R.~von~der Heydt.
\newblock A century of gestalt psychology in visual perception: I. perceptual
  grouping and figure--ground organization.
\newblock {\em Psychological bulletin}, 2012.

\bibitem{wilkes1992active}
D.~Wilkes and J.~K. Tsotsos.
\newblock Active object recognition.
\newblock In {\em Proceedings of the IEEE Conference on Computer Vision and
  Pattern Recognition (CVPR)}, 1992.

\bibitem{wu2018building}
Y.~Wu, Y.~Wu, G.~Gkioxari, and Y.~Tian.
\newblock Building generalizable agents with a realistic and rich {3D}
  environment.
\newblock {\em arXiv preprint arXiv:1801.02209}, 2018.

\bibitem{xiazamirhe2018gibsonenv}
F.~Xia, A.~R.~Zamir, Z.-Y. He, A.~Sax, J.~Malik, and S.~Savarese.
\newblock Gibson env: real-world perception for embodied agents.
\newblock In {\em Proceedings of the IEEE Conference on Computer Vision and
  Pattern Recognition (CVPR)}, 2018.

\bibitem{xiao2018video}
F.~Xiao and Y.~J. Lee.
\newblock Video object detection with an aligned spatial-temporal memory.
\newblock In {\em Proceedings of the European Conference on Computer Vision
  (ECCV)}, 2018.

\bibitem{ye2018active}
X.~Ye, Z.~Lin, H.~Li, S.~Zheng, and Y.~Yang.
\newblock Active object perceiver: Recognition-guided policy learning for
  object searching on mobile robots.
\newblock {\em IEEE/RSJ International Conference on Intelligent Robots and
  Systems (IROS)}, 2018.

\bibitem{yu2015multi}
F.~Yu and V.~Koltun.
\newblock Multi-scale context aggregation by dilated convolutions.
\newblock {\em International Conference on Lefvarning Representations (ICLR)},
  2015.

\bibitem{zhao2017pyramid}
H.~Zhao, J.~Shi, X.~Qi, X.~Wang, and J.~Jia.
\newblock Pyramid scene parsing network.
\newblock In {\em Proceedings of the IEEE Conference on Computer Vision and
  Pattern Recognition (CVPR)}, 2017.

\bibitem{zhou2017places}
B.~Zhou, A.~Lapedriza, A.~Khosla, A.~Oliva, and A.~Torralba.
\newblock Places: A 10 million image database for scene recognition.
\newblock {\em IEEE Transactions on Pattern Analysis and Machine Intelligence
  (TPAMI)}, 2017.

\bibitem{zhu2017target}
Y.~Zhu, R.~Mottaghi, E.~Kolve, J.~J. Lim, A.~Gupta, L.~Fei-Fei, and A.~Farhadi.
\newblock Target-driven visual navigation in indoor scenes using deep
  reinforcement learning.
\newblock In {\em IEEE International Conference on Robotics and Automation
  (ICRA)}, 2017.

\bibitem{zhu2017semantic}
Y.~Zhu, Y.~Tian, D.~Mexatas, and P.~Doll{\'a}r.
\newblock Semantic amodal segmentation.
\newblock In {\em Proceedings of the IEEE Conference on Computer Vision and
  Pattern Recognition (CVPR)}, 2017.

\end{thebibliography}
}

\clearpage
\onecolumn
\begin{appendices}

\section*{Appendix}
In the Appendix, we will provide more information about our dataset. 
\section{Object Category}

In our dataset, there are eight object categories, including bed, chair, desk, dresser, fridge, sofa, table, washer. In addition to Fig.4 in the main paper, we show the number of instances for each object category in Table~\ref{table:object_numbers}. As can be seen, the distribution over eight categories is fairly balanced. In Fig.~\ref{fig:category_ins}, we show some examples for each object category.

\begin{table*}[!ht]
\setlength{\tabcolsep}{3pt}

  \centering
  \begin{tabular}{l c c c c c c c c c c c c c}
    \toprule
      & bed & chair & desk & dresser & fridge & sofa & table & washer  & total \\
    \midrule	    
	\textbf{Train}  & 1687 & 1009 & 1333 & 737 & 900 & 1742 & 981 & 551 & 8940 \\
	\textbf{Val} & 197 & 122 & 207 & 82 & 103 & 206 & 144 & 52 & 1113 \\
    \textbf{Test} & 427 & 210 & 330 & 172 & 207 & 456 & 264 & 104 & 2170 \\
   \bottomrule
  \end{tabular}
    \caption{Number of instances for each object category.}
  \label{table:object_numbers}
\end{table*}

\begin{figure*}[!ht]
    \setlength{\tabcolsep}{1pt}
    \vspace{-4em}
    \begin{tabular}{ccccc}
    \includegraphics[width=0.2\linewidth]{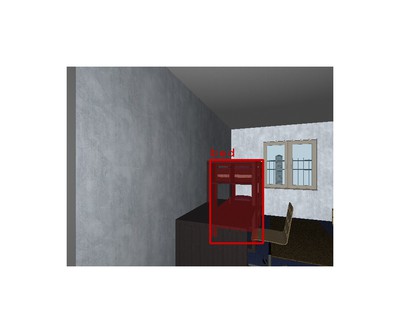}&
    \includegraphics[width=0.2\linewidth]{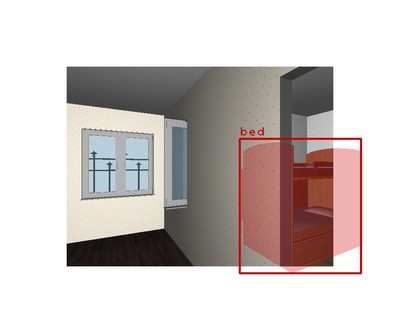}&
    \includegraphics[width=0.2\linewidth]{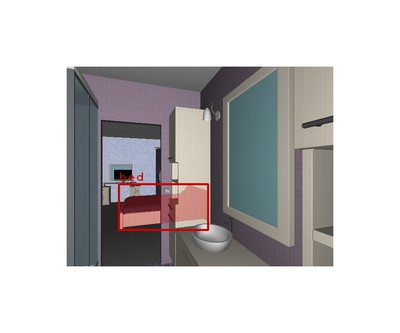}&
    \includegraphics[width=0.2\linewidth]{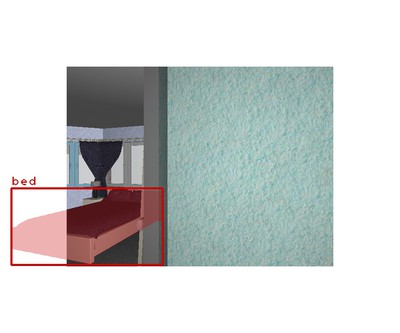}&
    \includegraphics[width=0.2\linewidth]{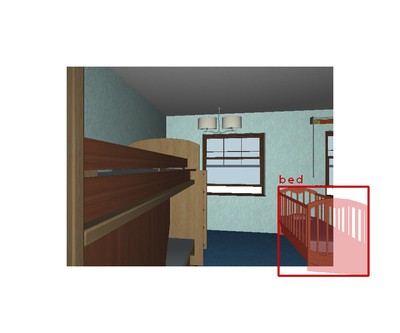} \\
    \includegraphics[width=0.2\linewidth]{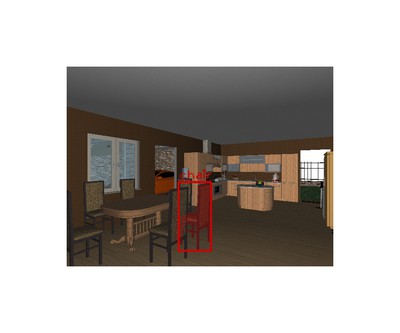}&
    \includegraphics[width=0.2\linewidth]{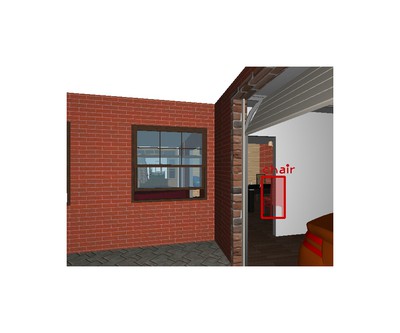}&
    \includegraphics[width=0.2\linewidth]{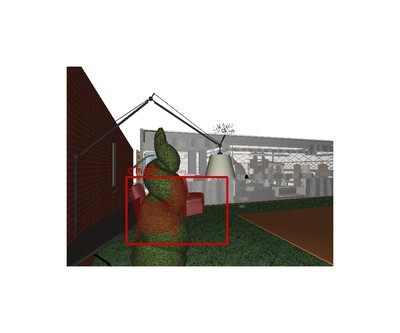}&
    \includegraphics[width=0.2\linewidth]{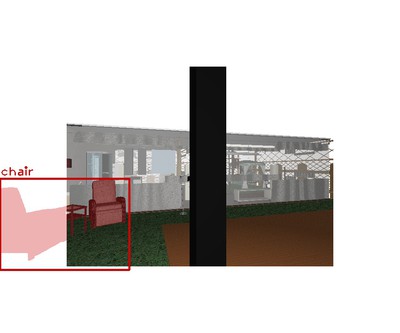}&
    \includegraphics[width=0.2\linewidth]{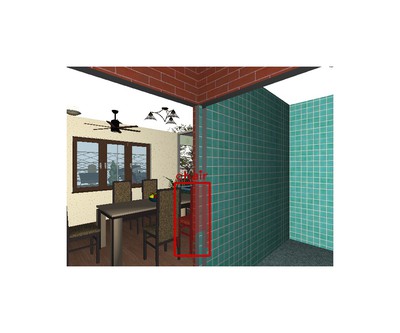} \\
    \includegraphics[width=0.2\linewidth]{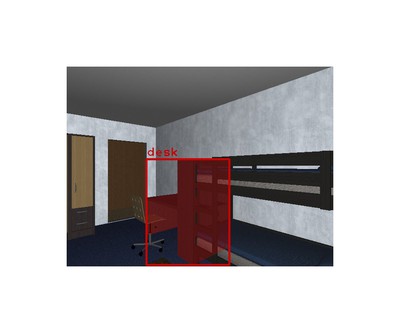}&
    \includegraphics[width=0.2\linewidth]{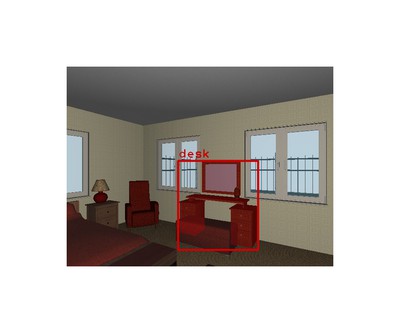}&
    \includegraphics[width=0.2\linewidth]{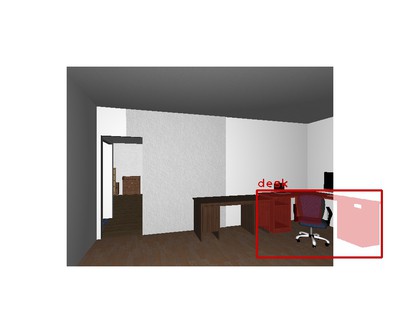}&
    \includegraphics[width=0.2\linewidth]{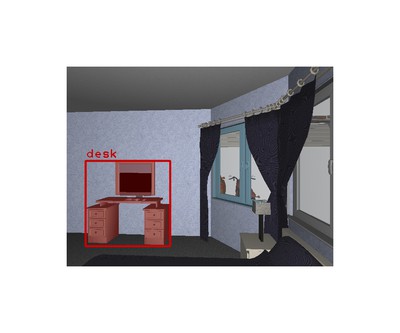}&
    \includegraphics[width=0.2\linewidth]{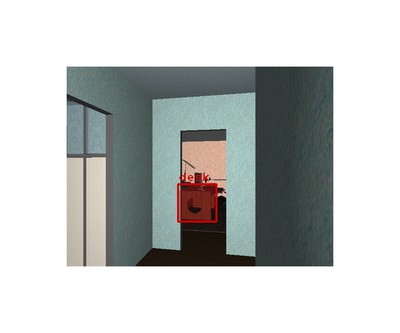} \\
    \includegraphics[width=0.2\linewidth]{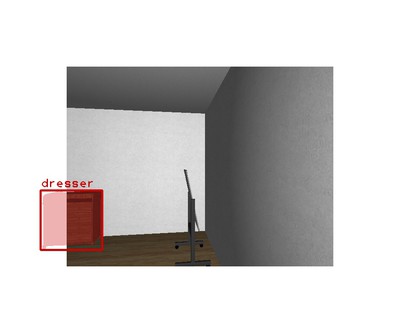}&
    \includegraphics[width=0.2\linewidth]{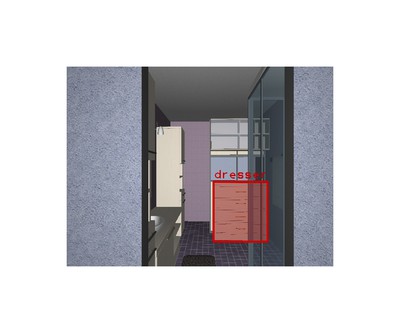}&
    \includegraphics[width=0.2\linewidth]{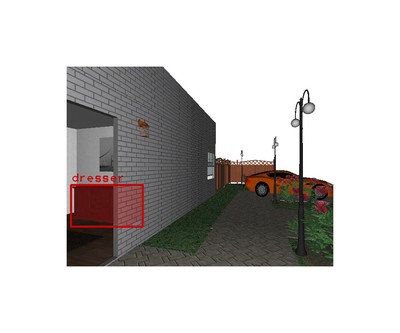}&
    \includegraphics[width=0.2\linewidth]{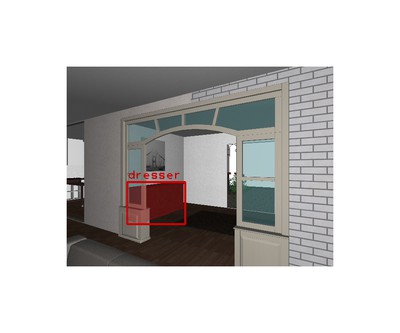}&
    \includegraphics[width=0.2\linewidth]{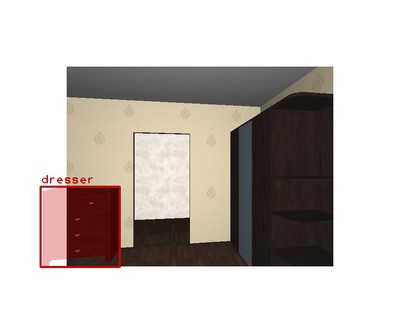} \\
    \includegraphics[width=0.2\linewidth]{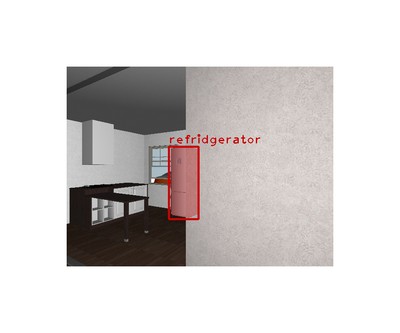}&
    \includegraphics[width=0.2\linewidth]{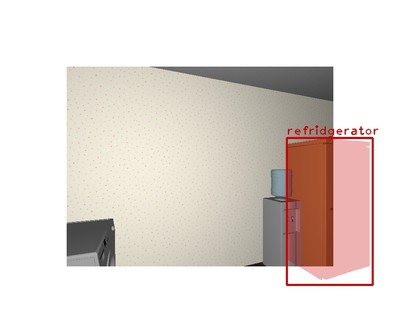}&
    \includegraphics[width=0.2\linewidth]{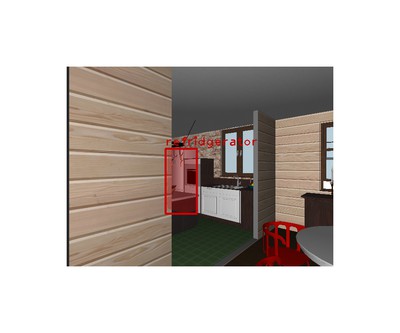}&
    \includegraphics[width=0.2\linewidth]{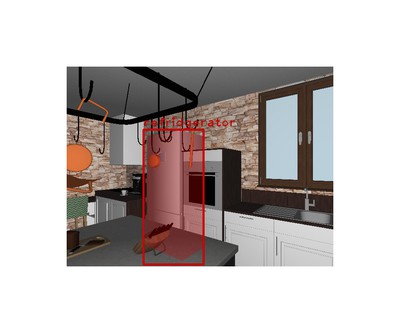}&
    \includegraphics[width=0.2\linewidth]{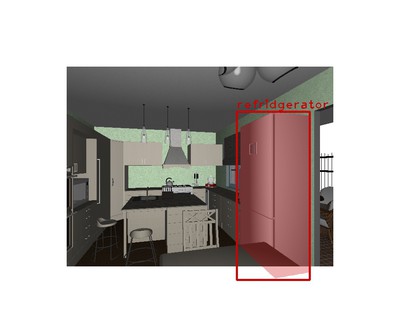} \\
    \includegraphics[width=0.2\linewidth]{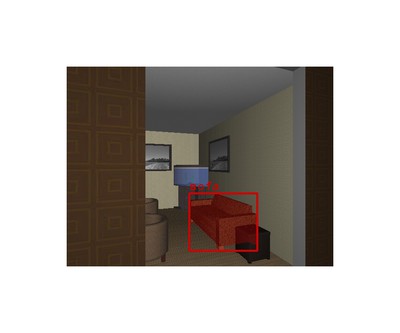}&
    \includegraphics[width=0.2\linewidth]{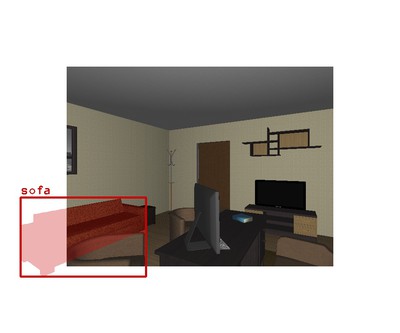}&
    \includegraphics[width=0.2\linewidth]{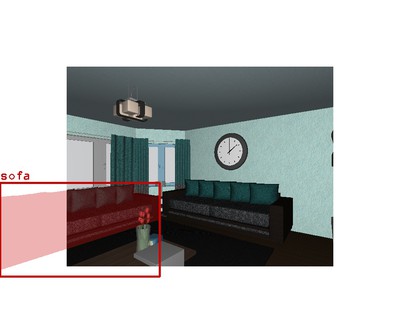}&
    \includegraphics[width=0.2\linewidth]{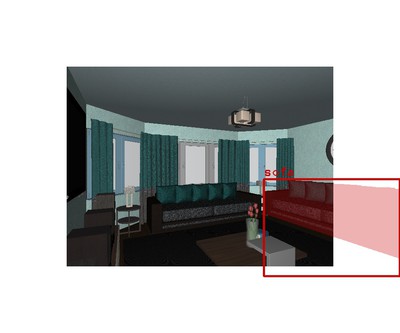}&
    \includegraphics[width=0.2\linewidth]{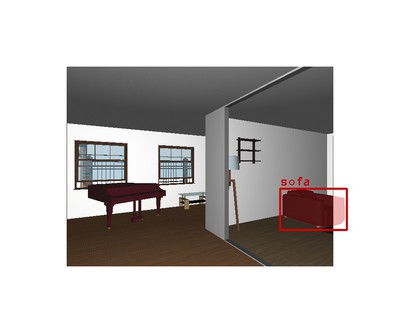} \\
    \includegraphics[width=0.2\linewidth]{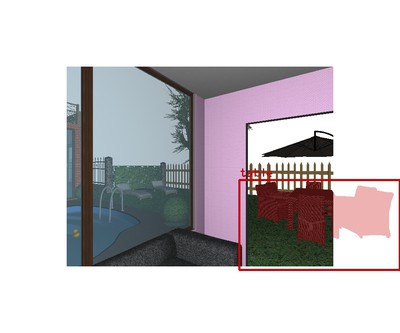}&
    \includegraphics[width=0.2\linewidth]{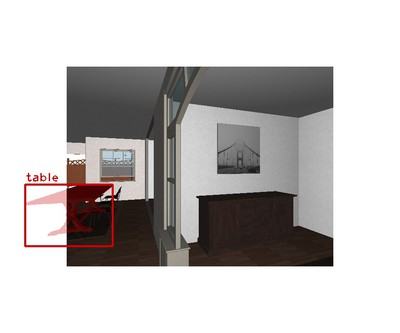}&
    \includegraphics[width=0.2\linewidth]{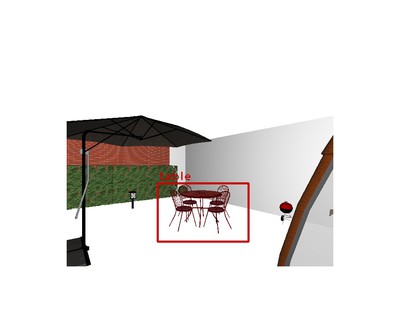}&
    \includegraphics[width=0.2\linewidth]{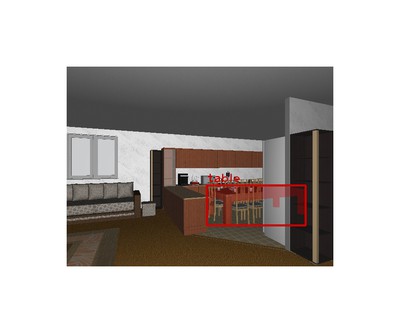}&
    \includegraphics[width=0.2\linewidth]{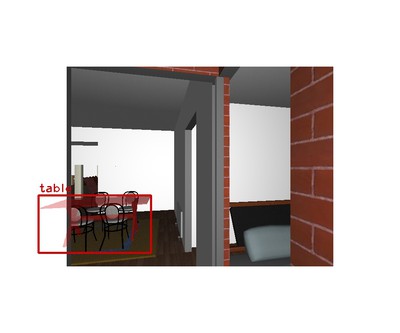} \\
    \includegraphics[width=0.2\linewidth]{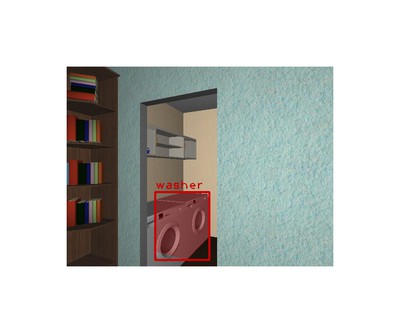}&
    \includegraphics[width=0.2\linewidth]{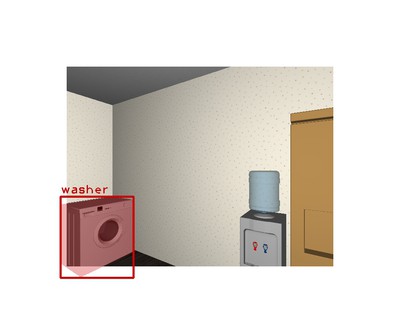}&
    \includegraphics[width=0.2\linewidth]{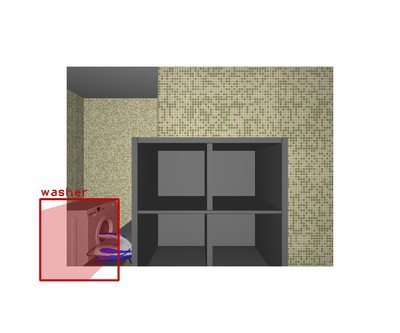}&
    \includegraphics[width=0.2\linewidth]{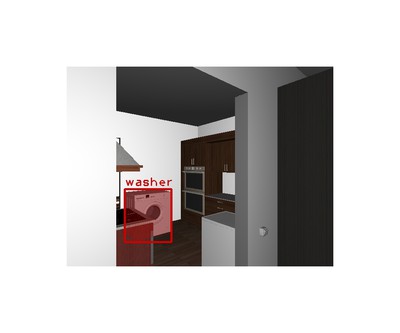}&
    \includegraphics[width=0.2\linewidth]{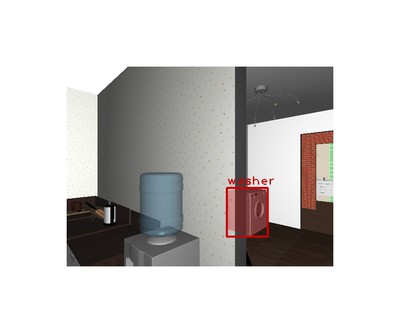} \\    
    \end{tabular}
    \caption{Visualizing examples of our dataset. In each row, we visualize ground-truth annotations for bed, chair, desk, dresser, fridge, sofa, table, washer.}
    \label{fig:category_ins}
\end{figure*}

\section{Shortest Path}
\begin{figure*}[!ht]
    \centering
    \begin{tabular}{c}
   	\includegraphics[width=0.24\linewidth, trim={15cm 4.8cm 0 0cm},clip]{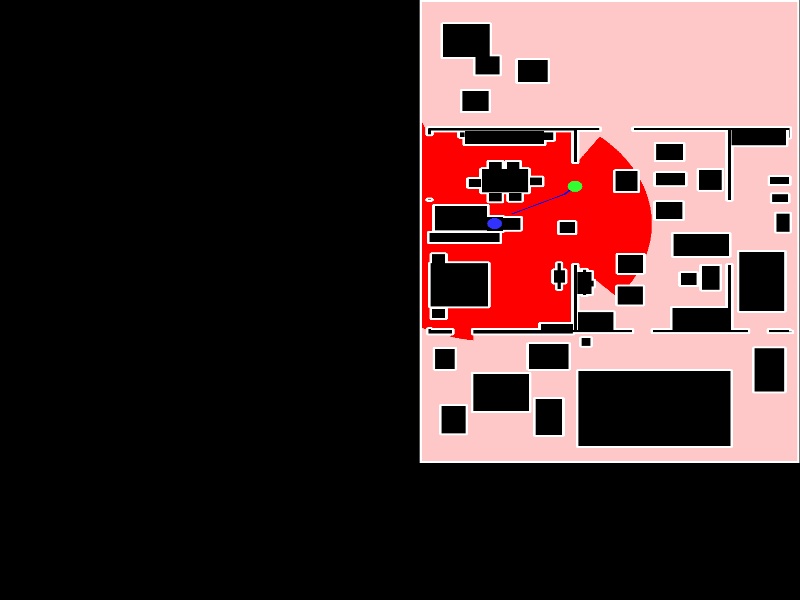}
   	\includegraphics[width=0.24\linewidth, trim={15cm 4.8cm 0 0cm},clip]{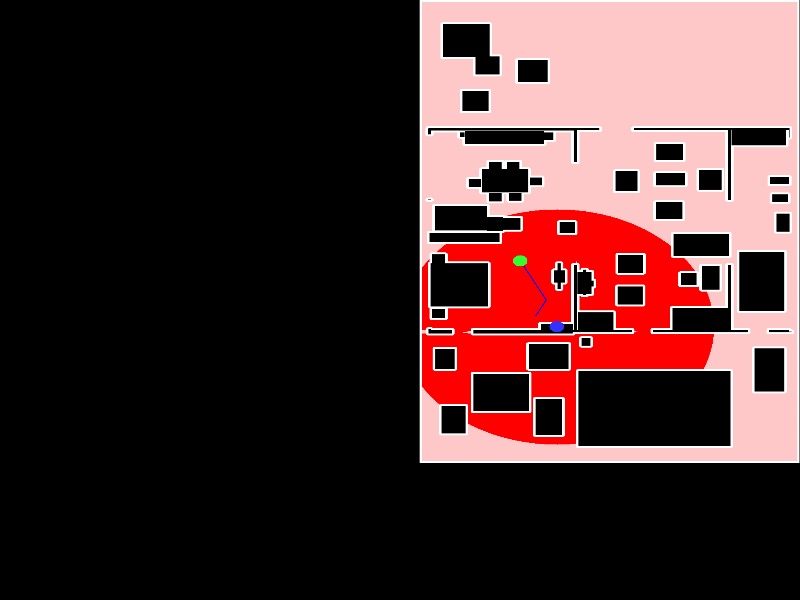}
   	\includegraphics[width=0.24\linewidth, trim={15cm 4.8cm 0 0cm},clip]{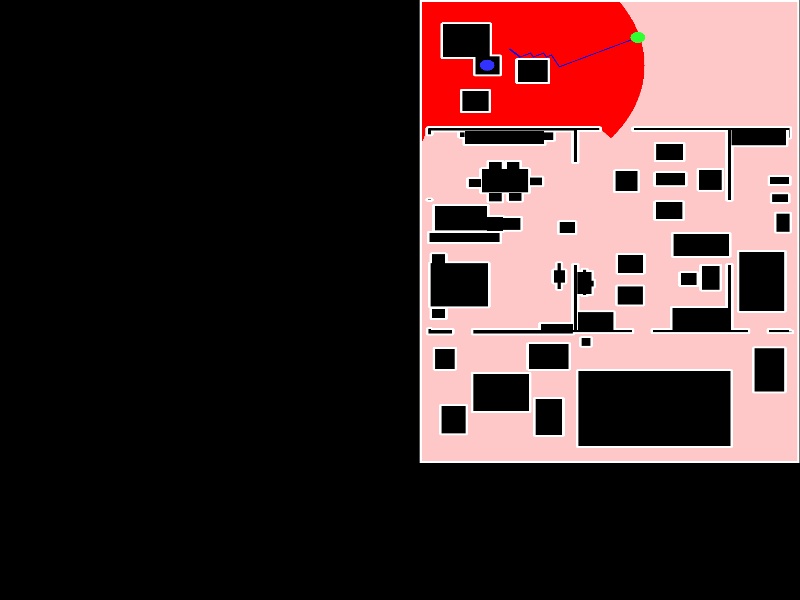}
   	\includegraphics[width=0.24\linewidth, trim={15cm 4.8cm 0 0cm},clip]{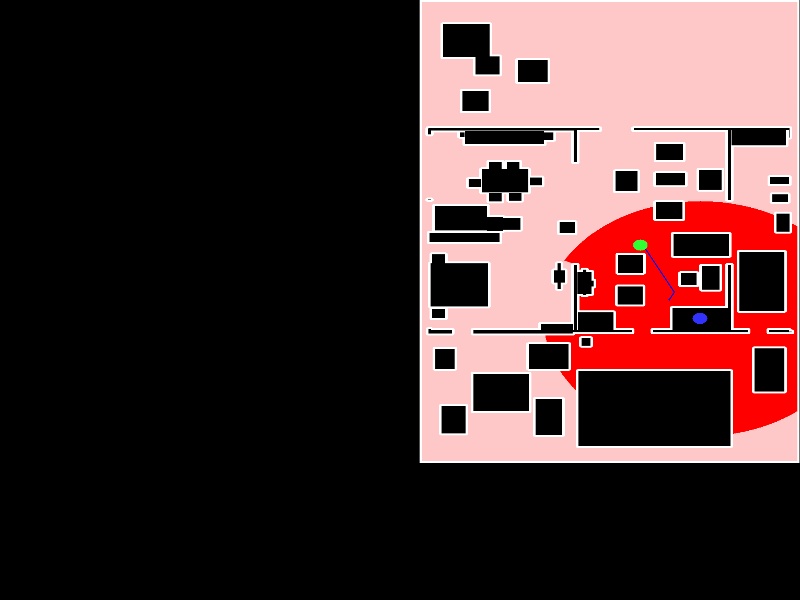}   	   \\ 
   	\includegraphics[width=0.24\linewidth, trim={7cm 7cm 7cm 7cm},clip]{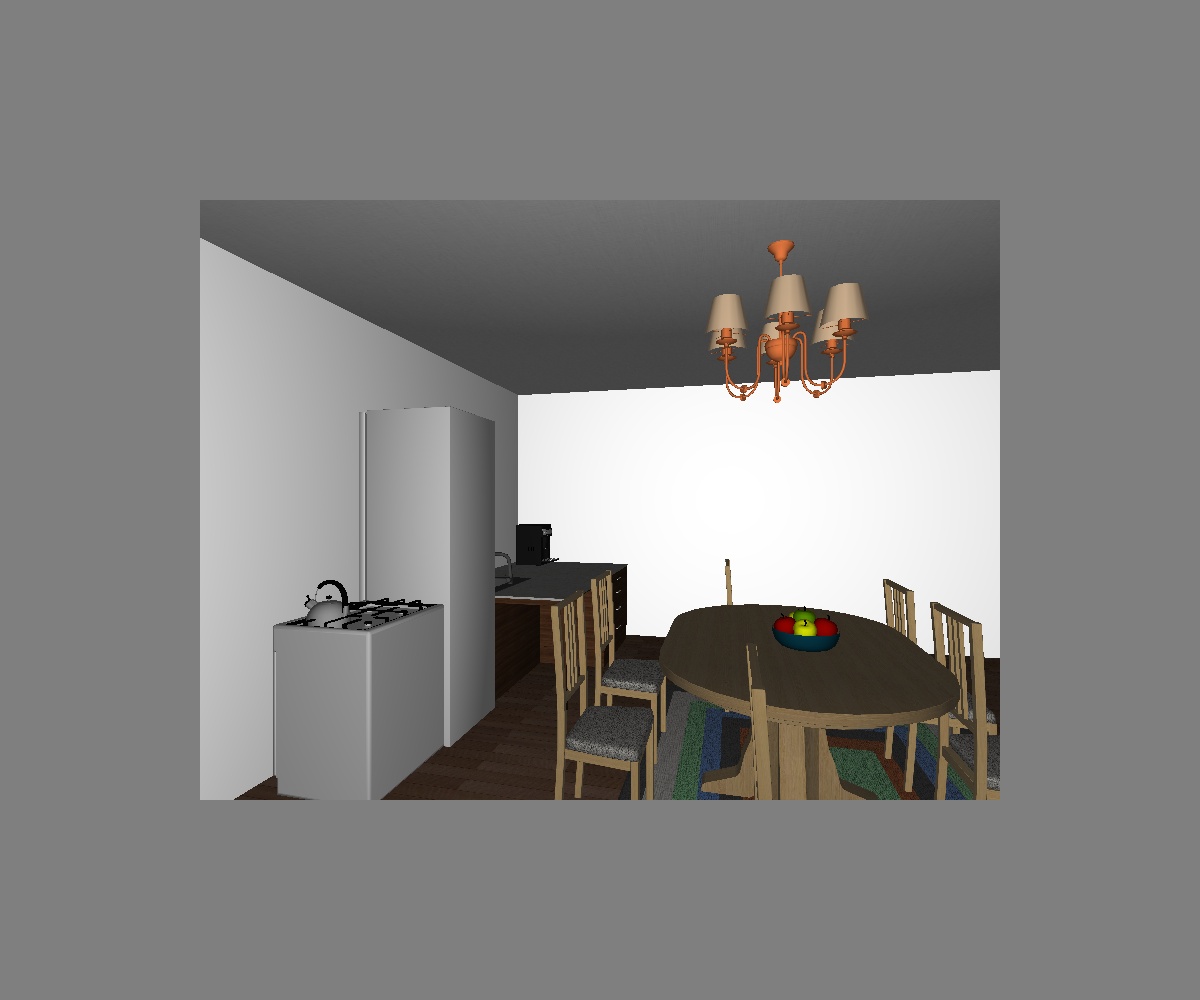}
   	\includegraphics[width=0.24\linewidth, trim={7cm 7cm 7cm 7cm},clip]{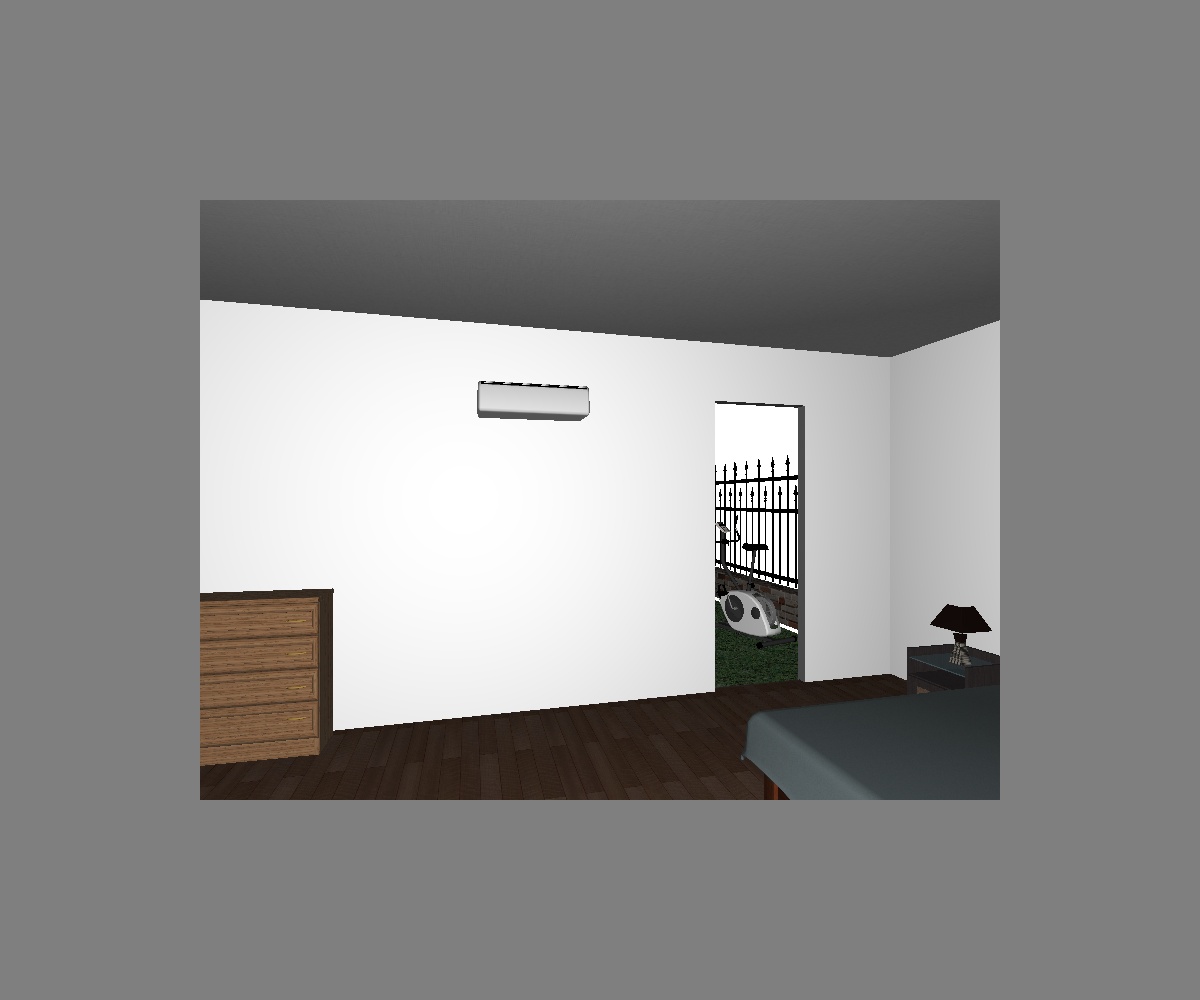}
   	\includegraphics[width=0.24\linewidth, trim={7cm 7cm 7cm 7cm},clip]{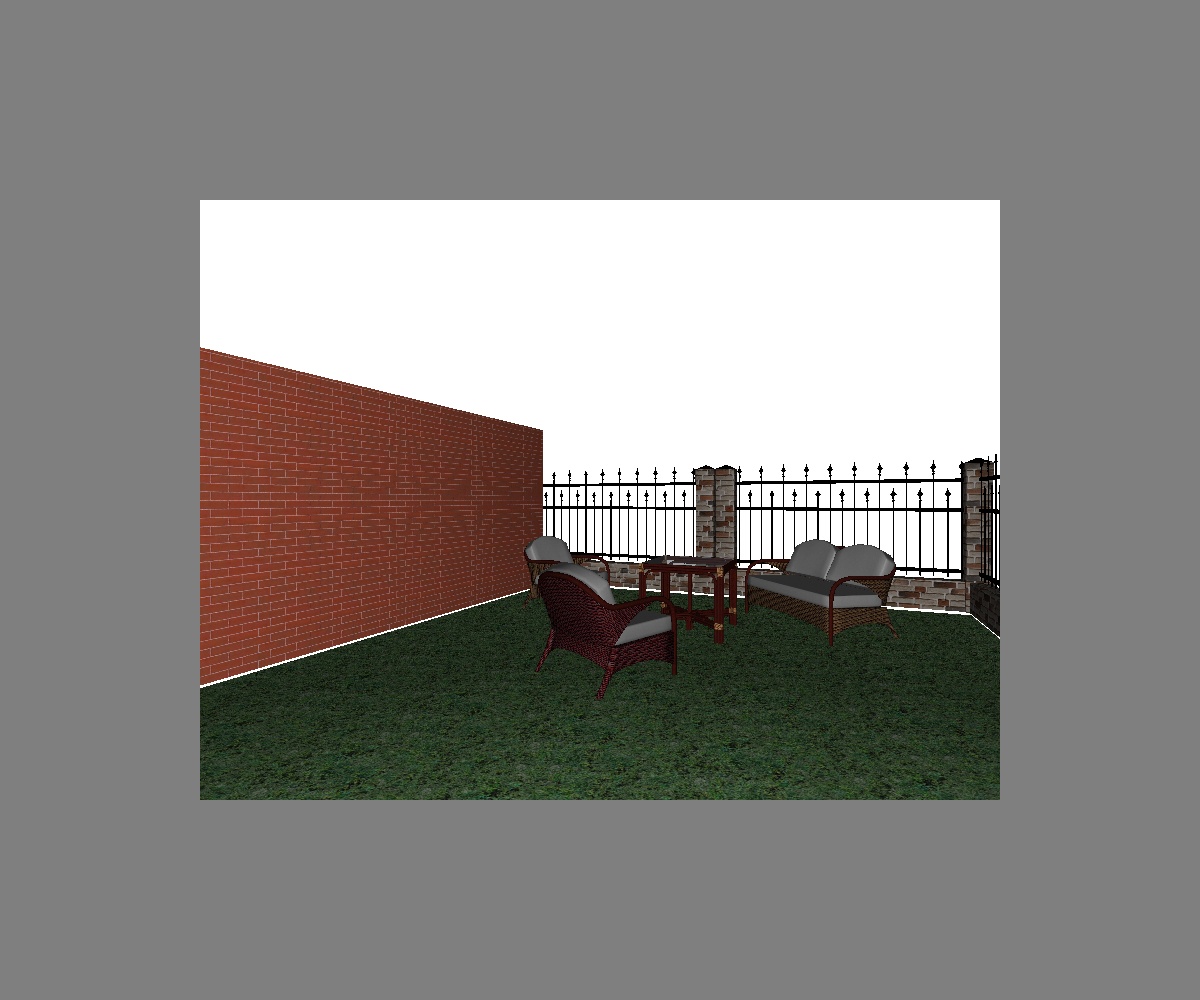}
   	\includegraphics[width=0.24\linewidth, trim={7cm 7cm 7cm 7cm},clip]{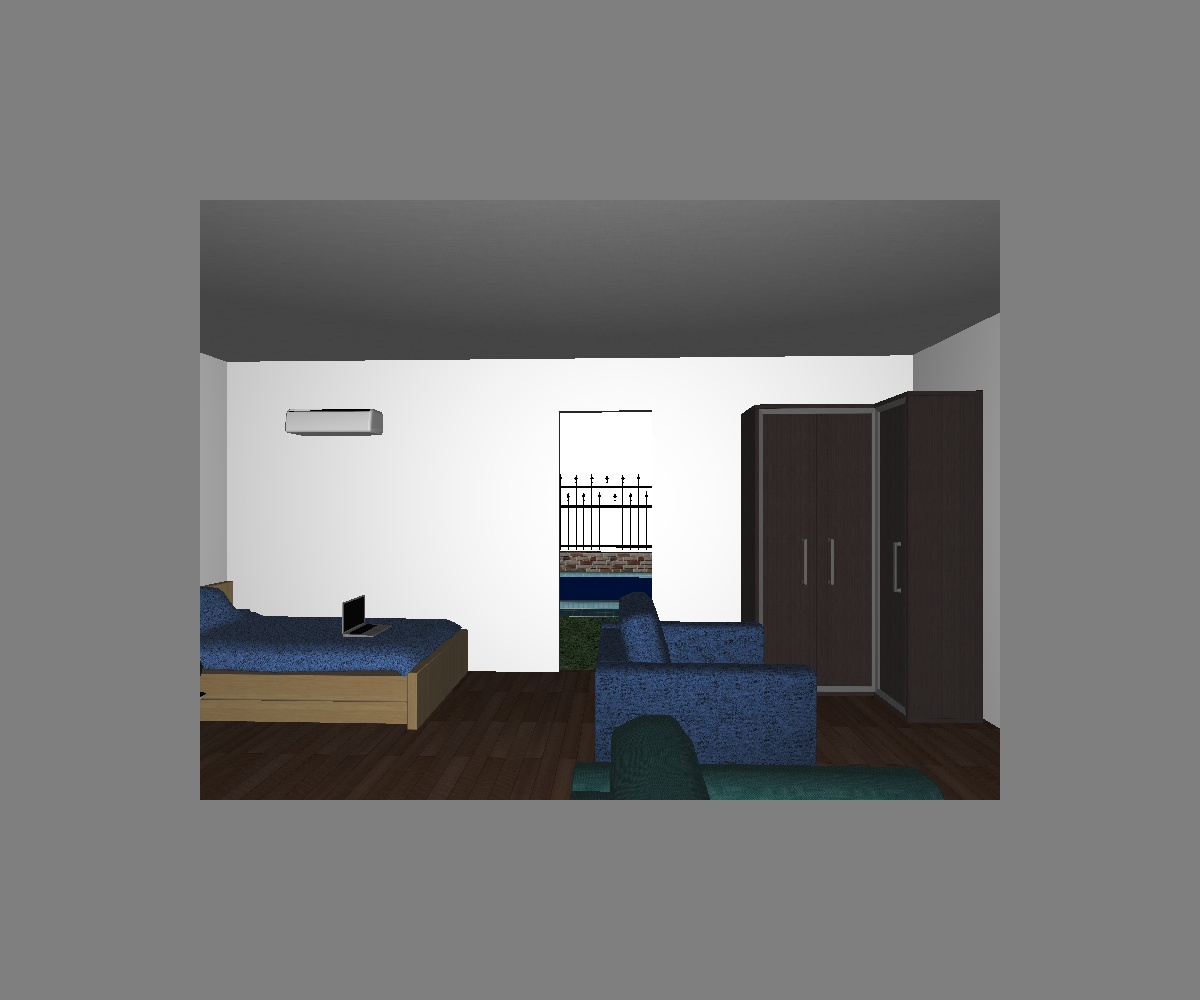} \\
   	\includegraphics[width=0.24\linewidth, trim={7cm 7cm 7cm 7cm},clip]{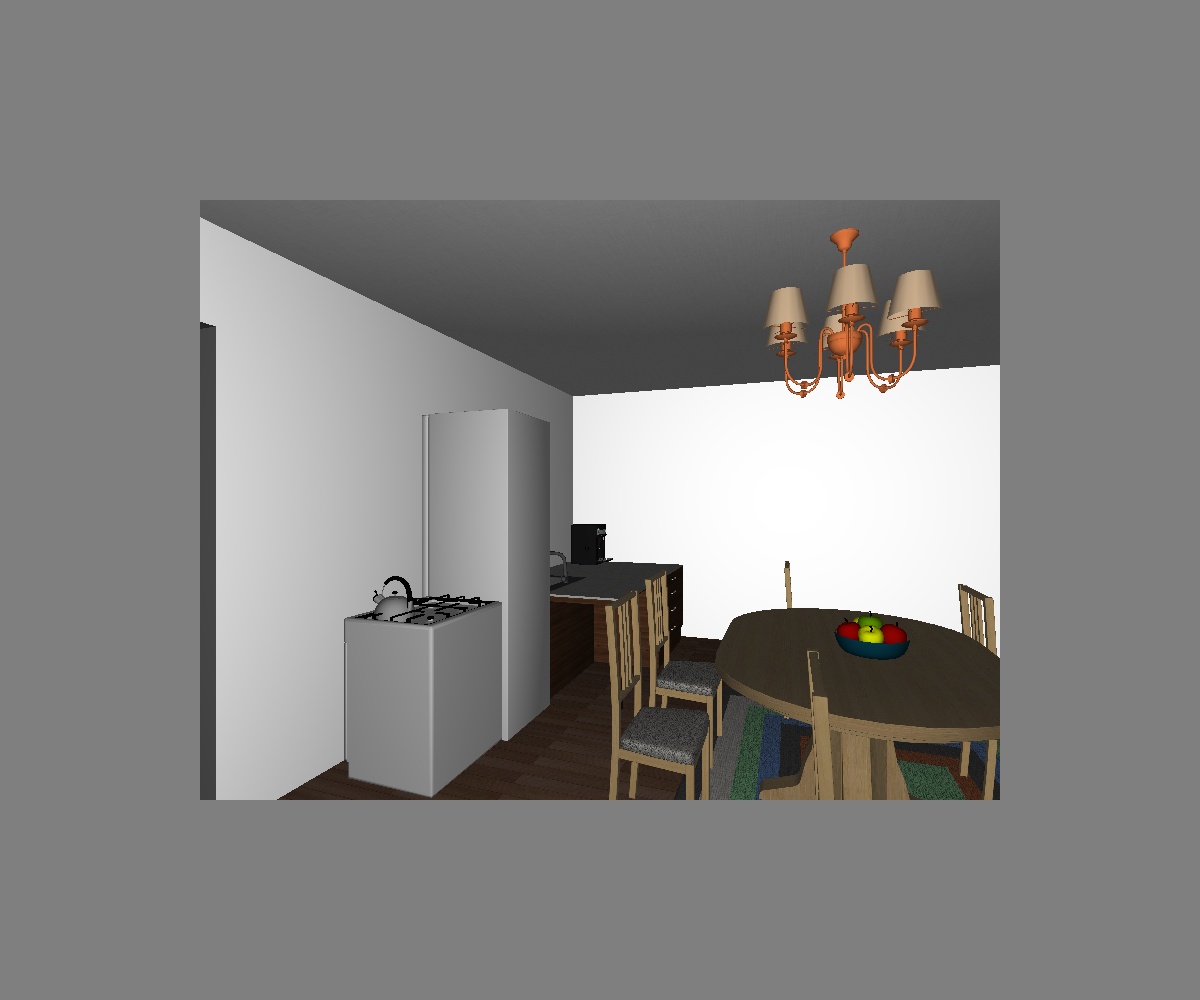}
   	\includegraphics[width=0.24\linewidth, trim={7cm 7cm 7cm 7cm},clip]{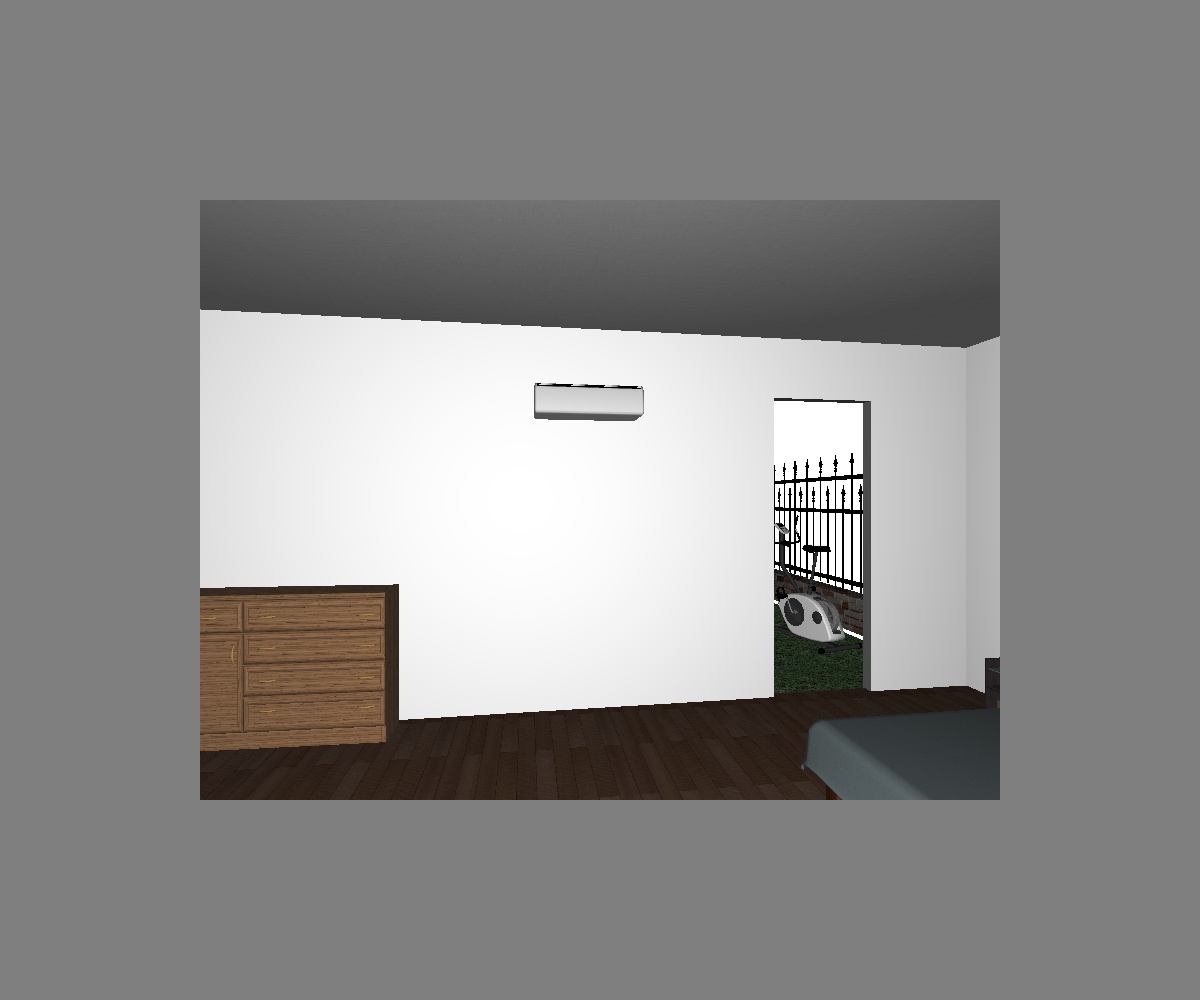}
   	\includegraphics[width=0.24\linewidth, trim={7cm 7cm 7cm 7cm},clip]{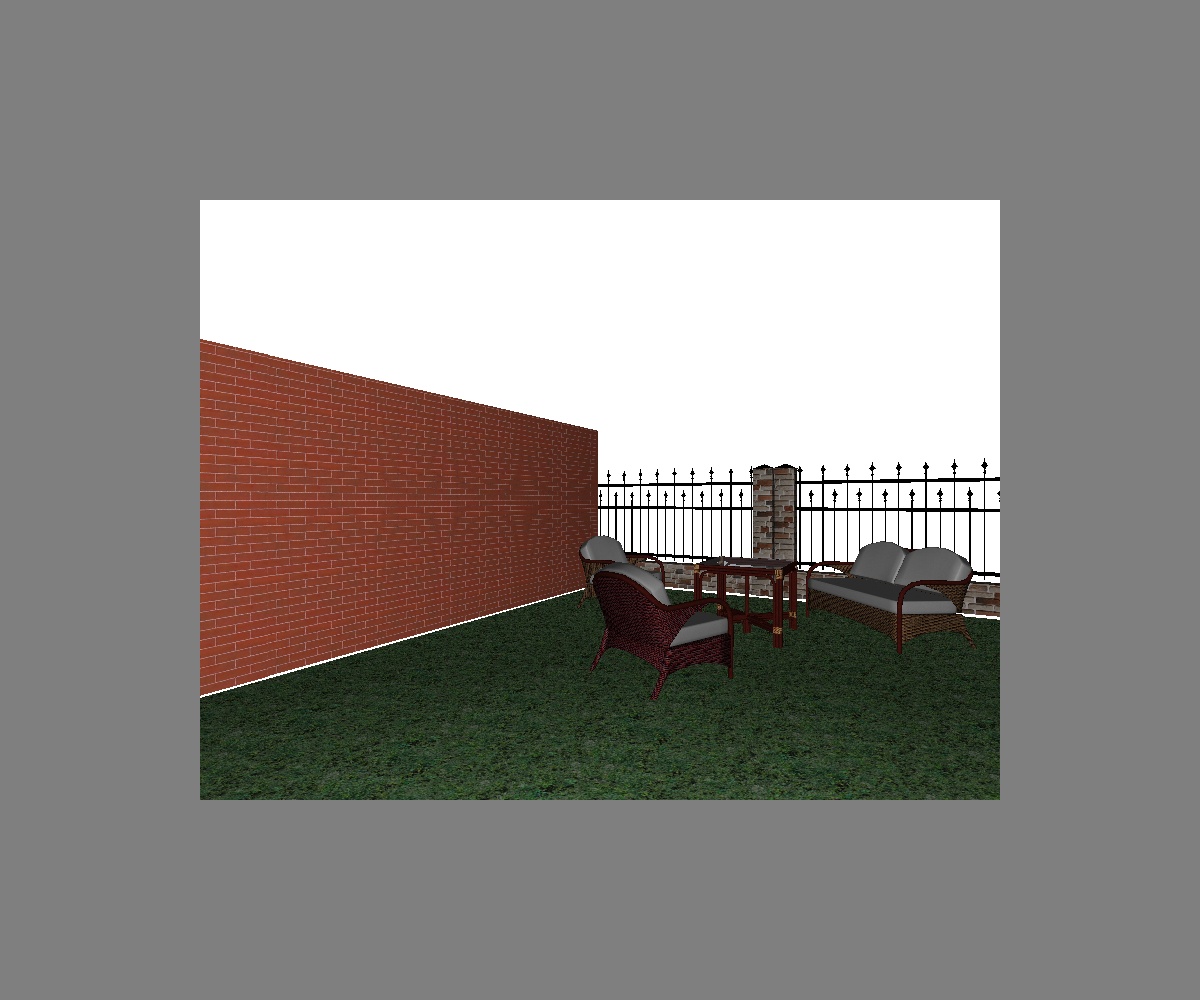}
   	\includegraphics[width=0.24\linewidth, trim={7cm 7cm 7cm 7cm},clip]{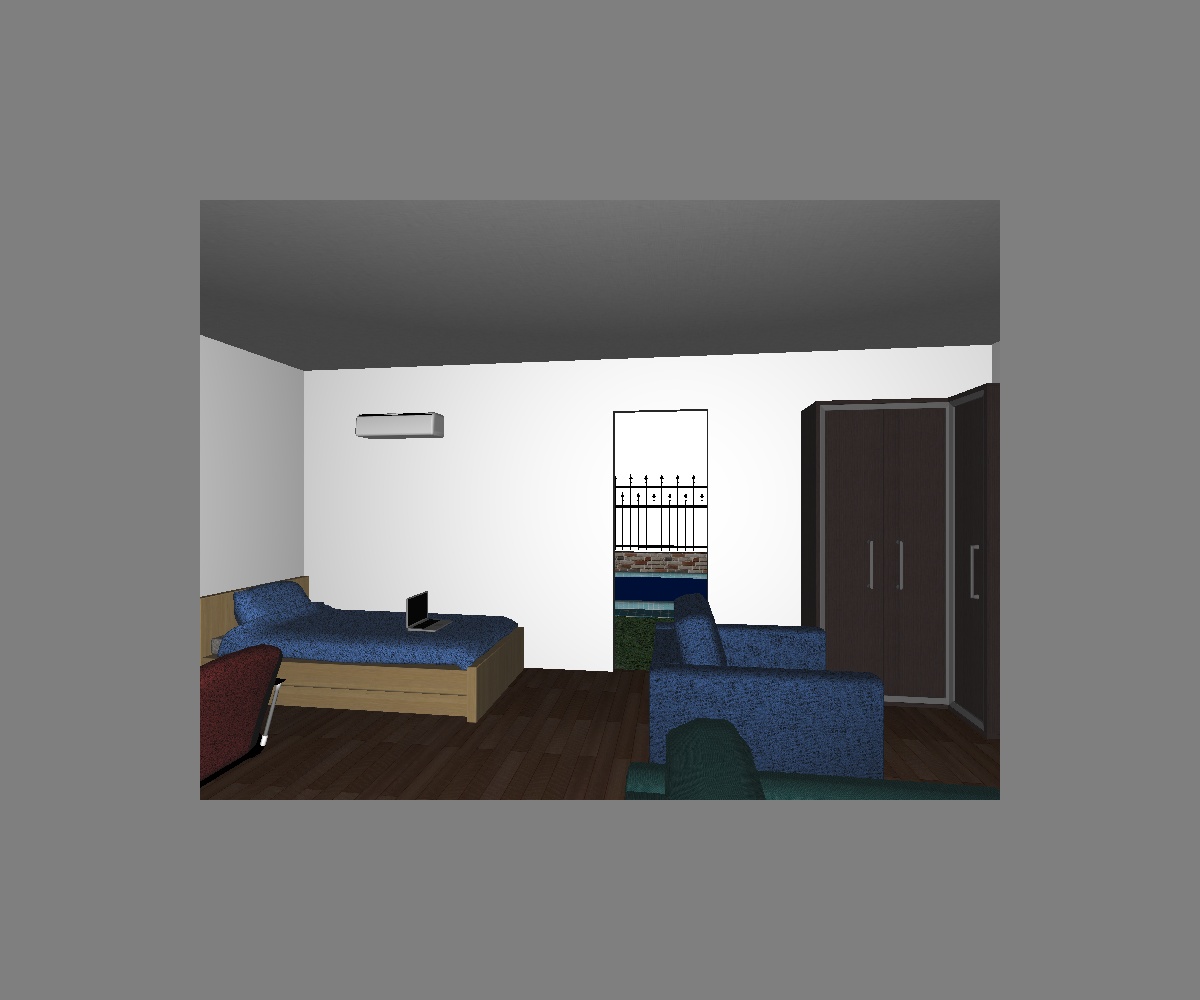}    \\
   	\includegraphics[width=0.24\linewidth, trim={7cm 7cm 7cm 7cm},clip]{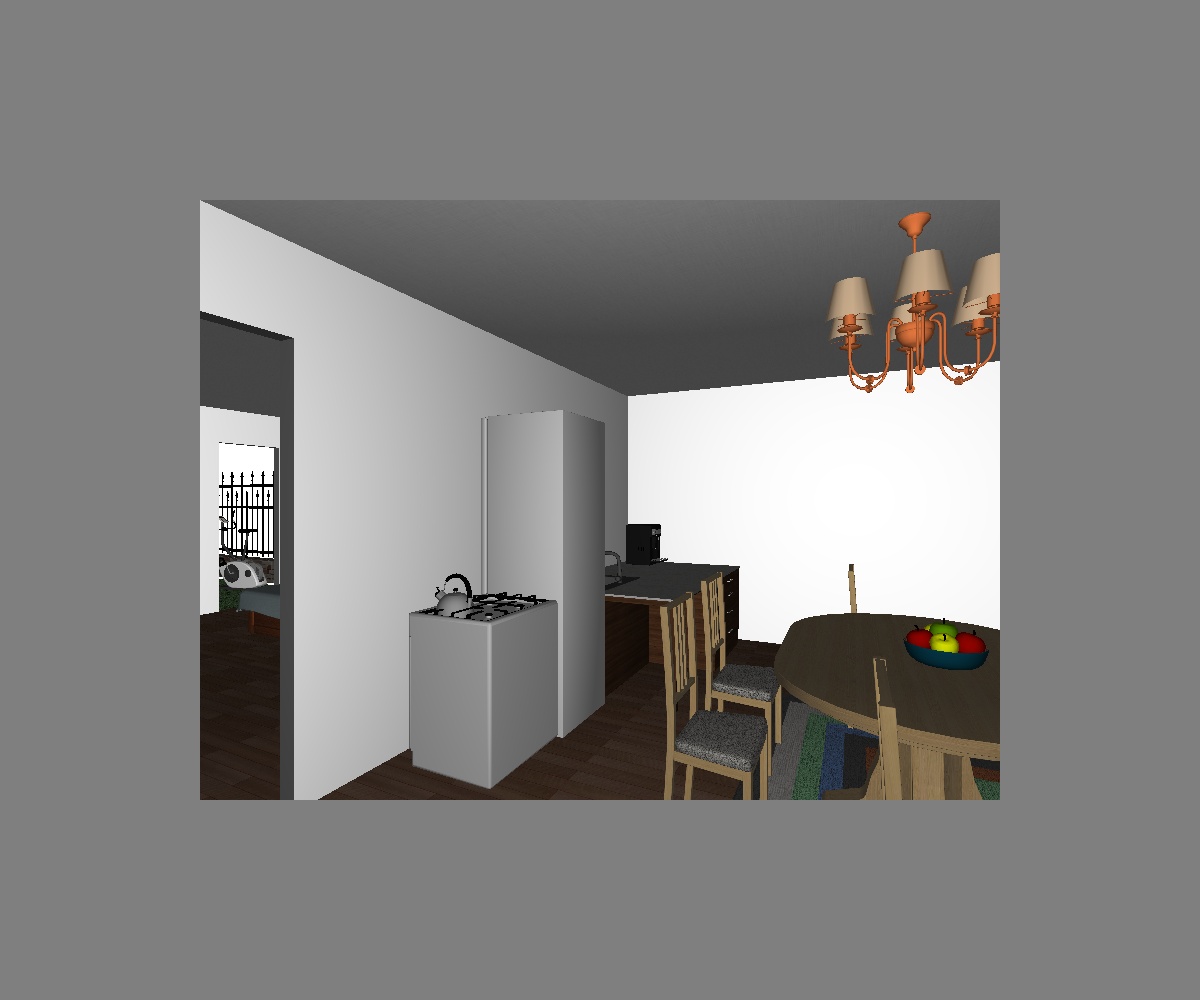}
   	\includegraphics[width=0.24\linewidth, trim={7cm 7cm 7cm 7cm},clip]{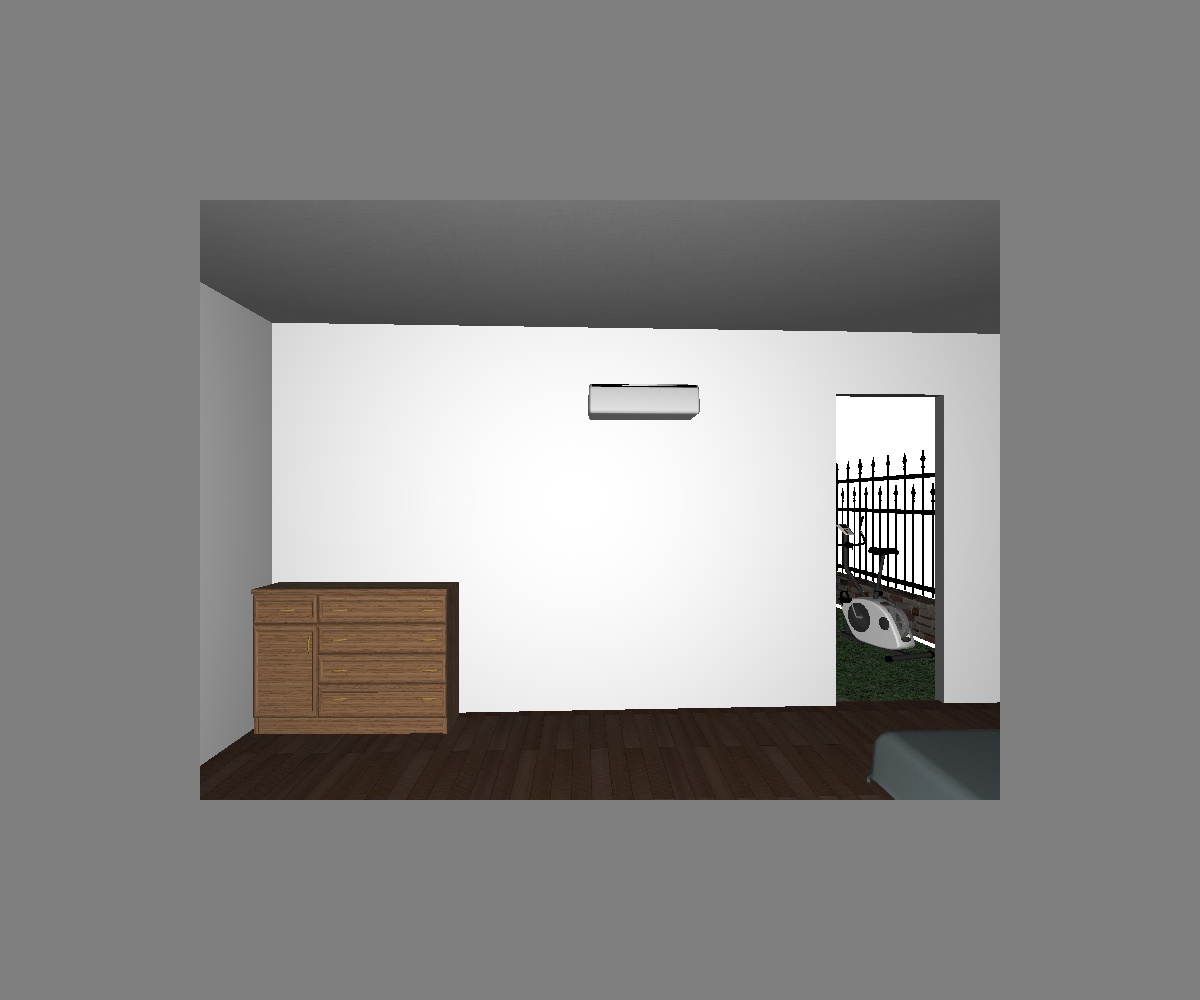}
   	\includegraphics[width=0.24\linewidth, trim={7cm 7cm 7cm 7cm},clip]{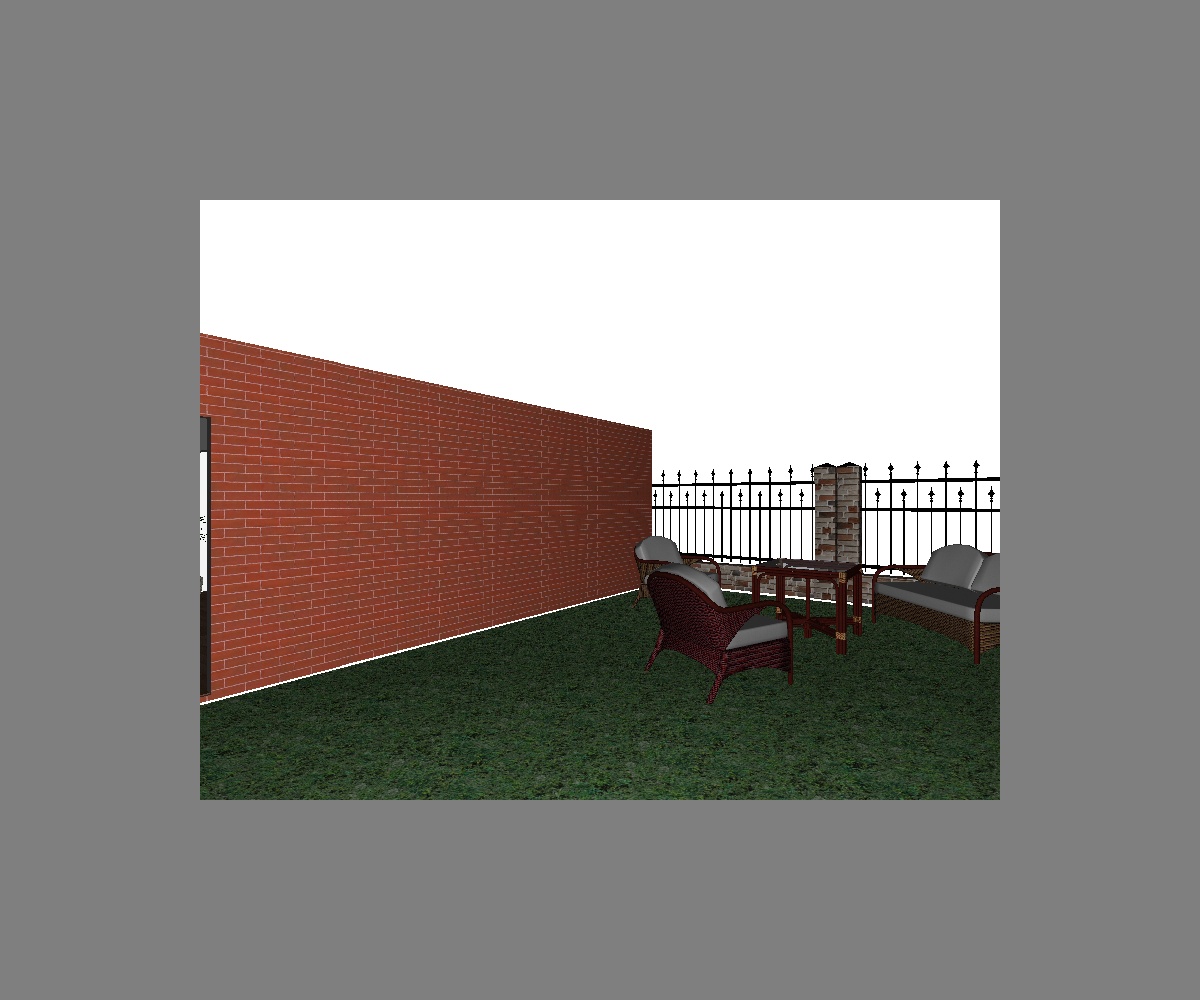}
   	\includegraphics[width=0.24\linewidth, trim={7cm 7cm 7cm 7cm},clip]{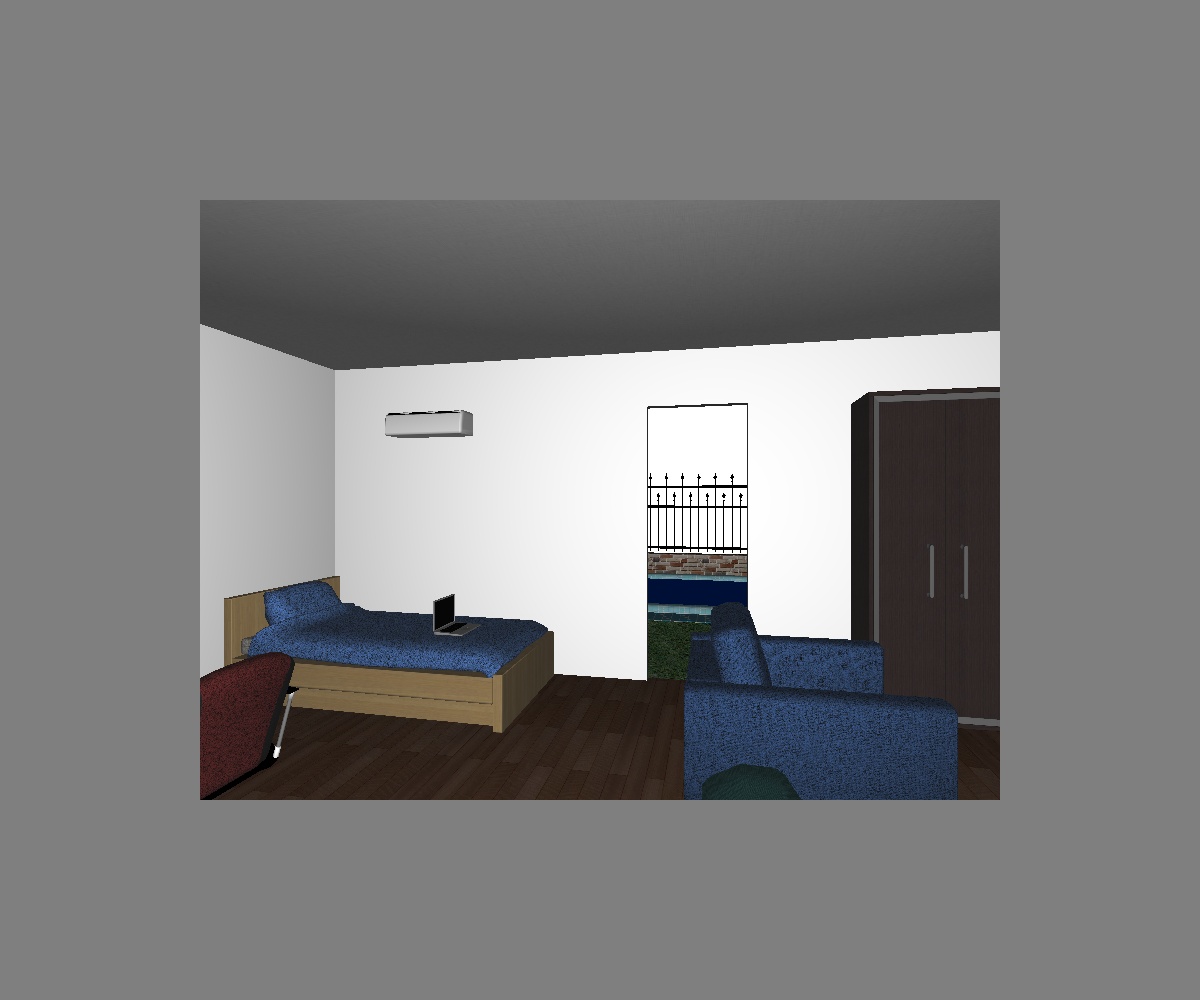}   	\\
   	\includegraphics[width=0.24\linewidth, trim={7cm 7cm 7cm 7cm},clip]{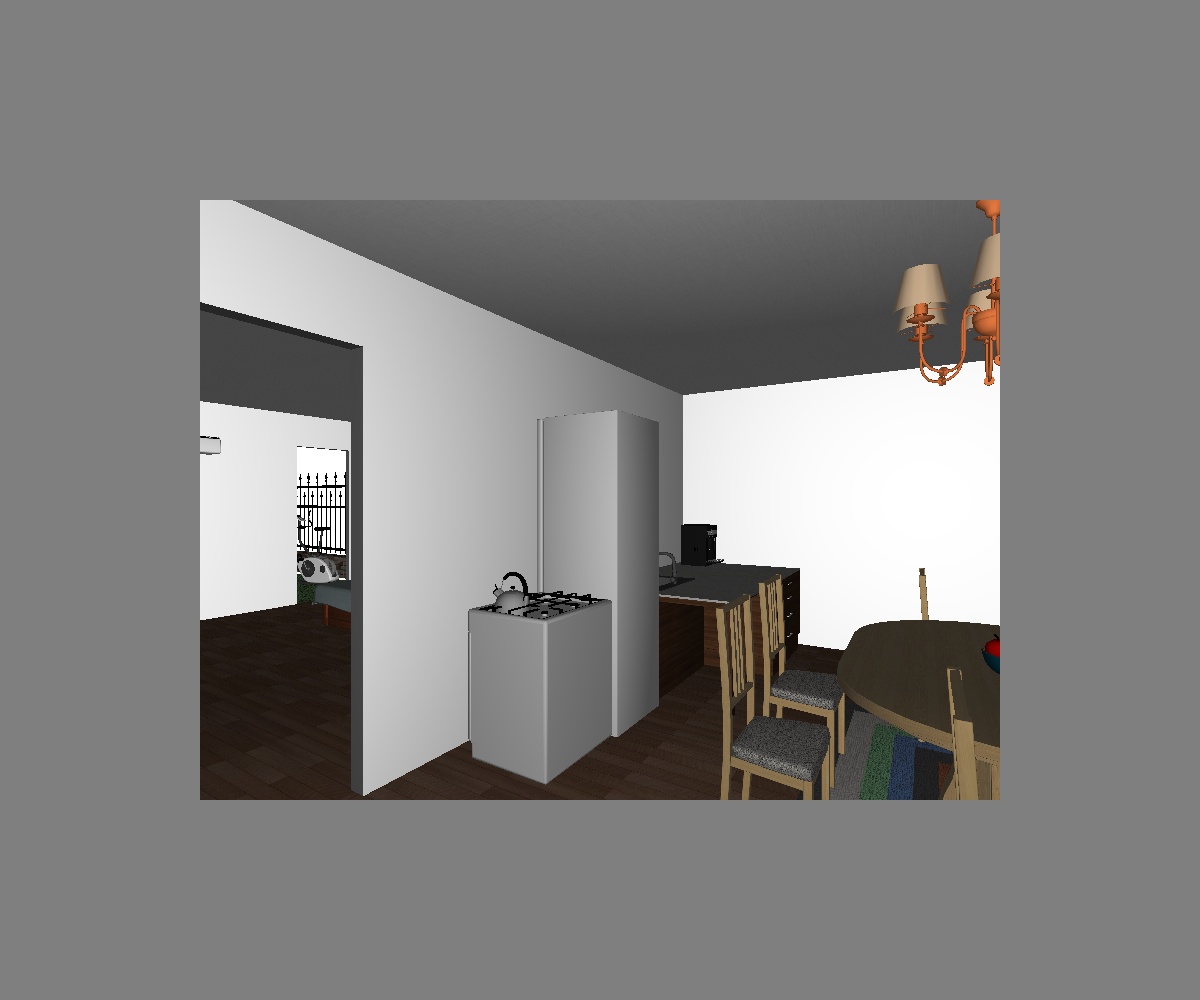}
   	\includegraphics[width=0.24\linewidth, trim={7cm 7cm 7cm 7cm},clip]{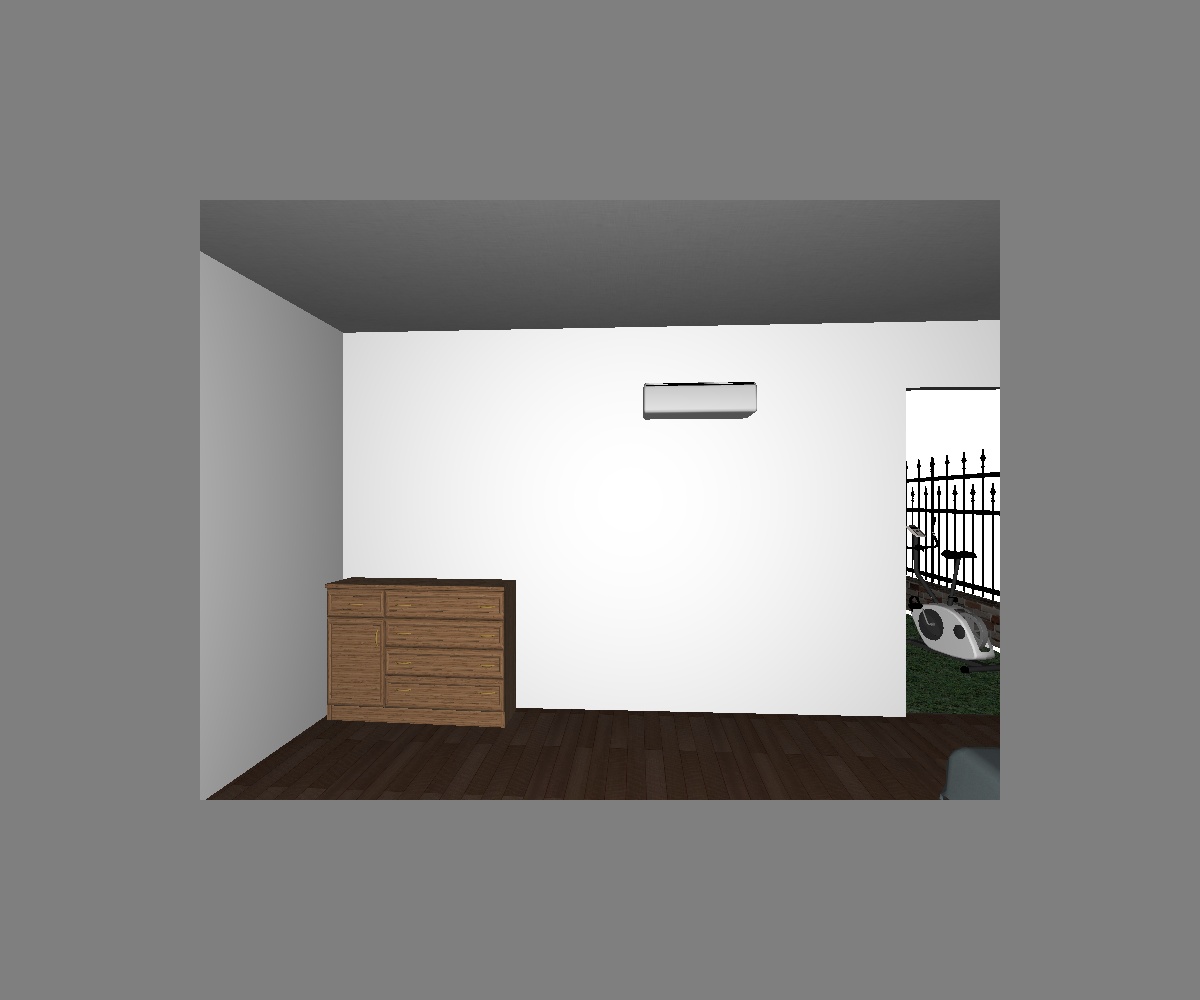}
   	\includegraphics[width=0.24\linewidth, trim={7cm 7cm 7cm 7cm},clip]{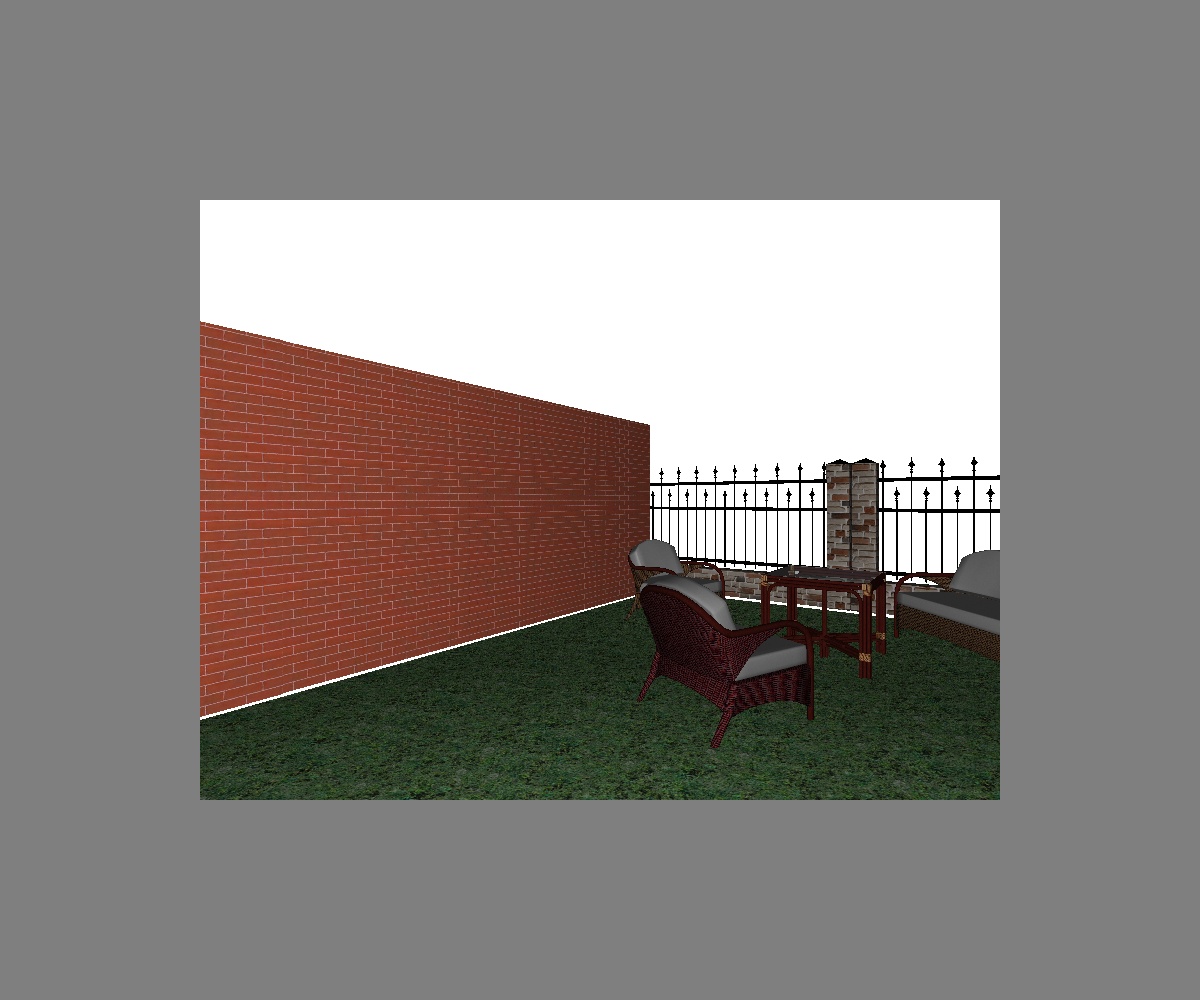}
   	\includegraphics[width=0.24\linewidth, trim={7cm 7cm 7cm 7cm},clip]{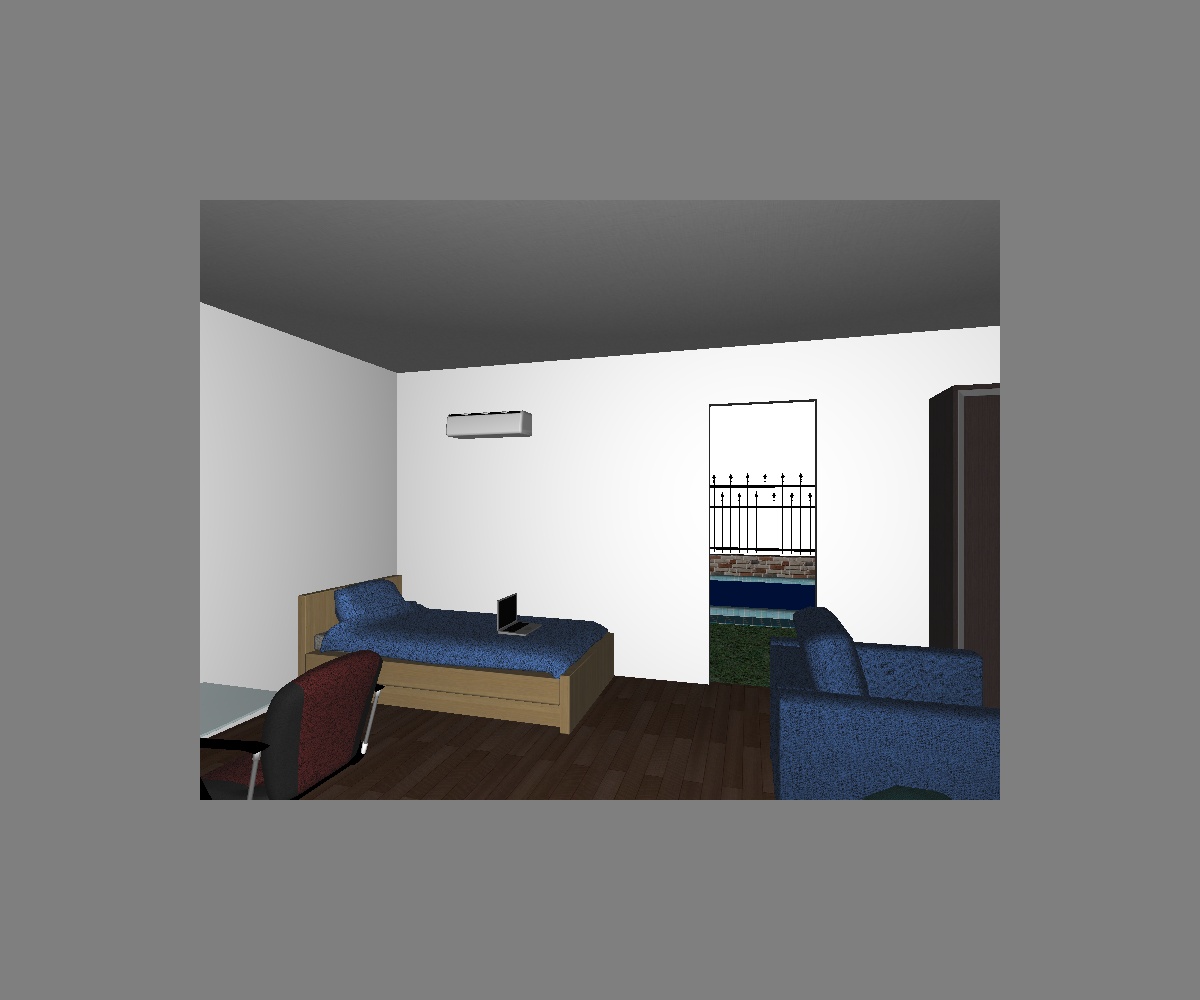}   	\\
   	\includegraphics[width=0.24\linewidth, trim={7cm 7cm 7cm 7cm},clip]{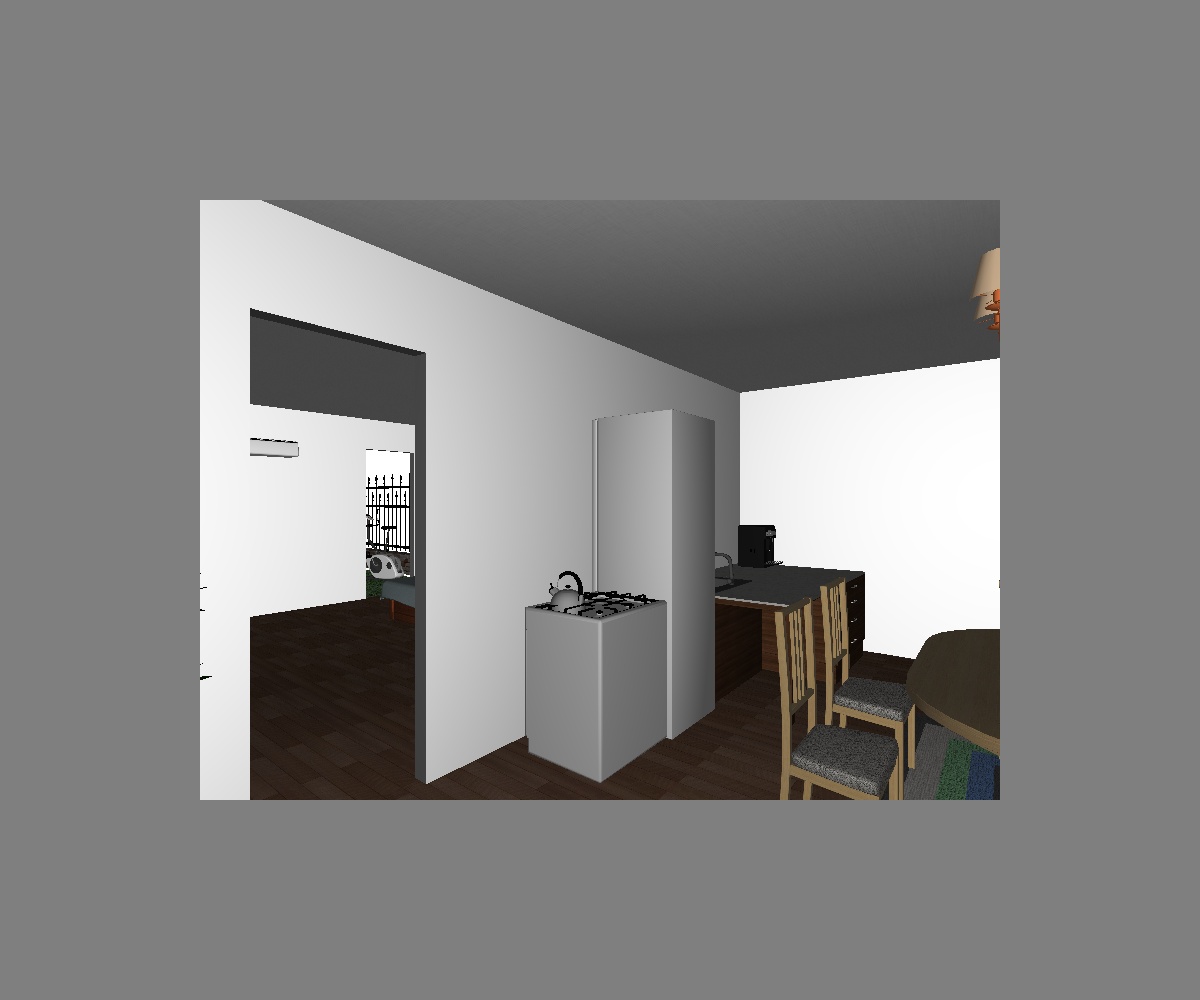}
   	\includegraphics[width=0.24\linewidth, trim={7cm 7cm 7cm 7cm},clip]{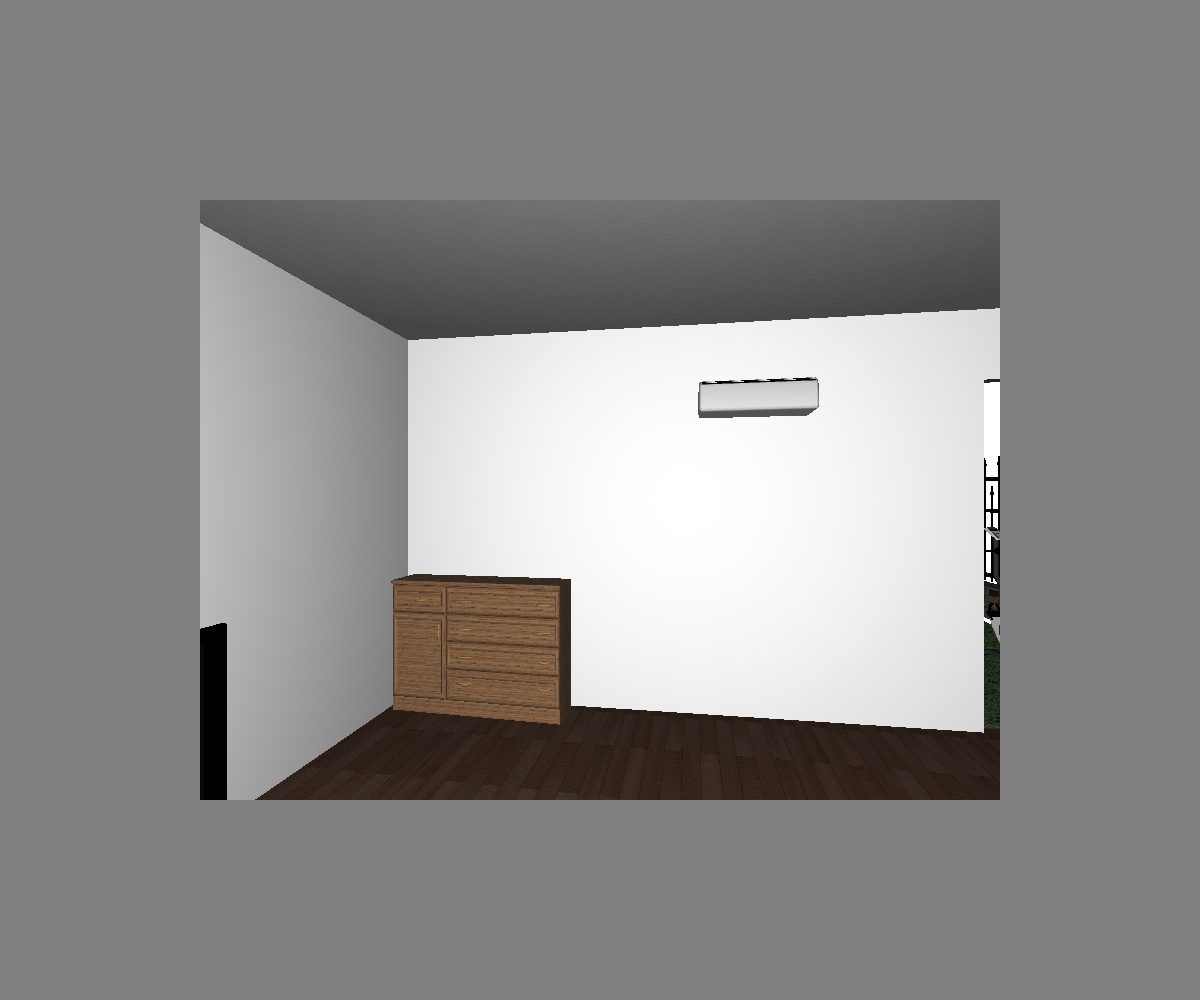}
   	\includegraphics[width=0.24\linewidth, trim={7cm 7cm 7cm 7cm},clip]{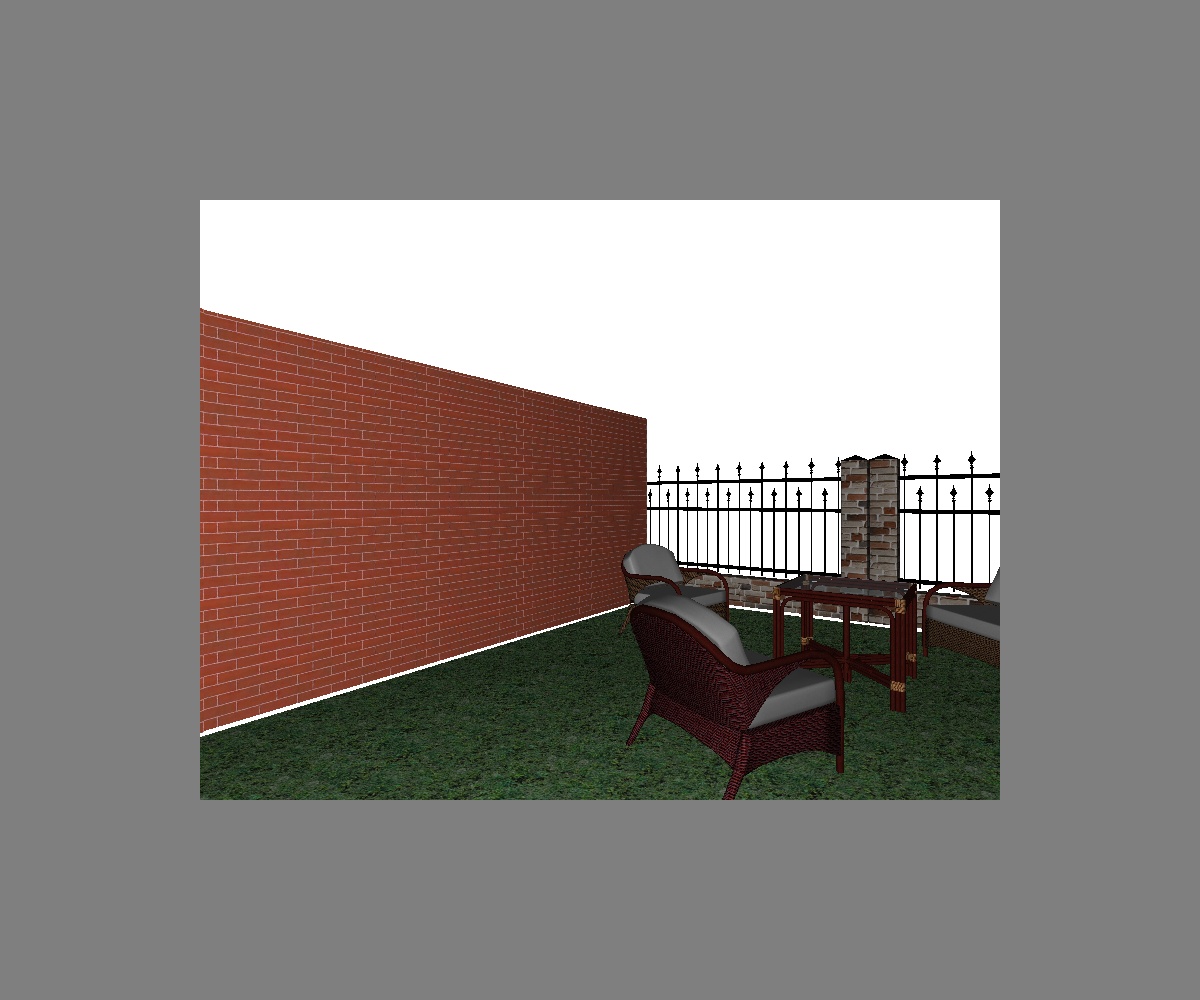}
   	\includegraphics[width=0.24\linewidth, trim={7cm 7cm 7cm 7cm},clip]{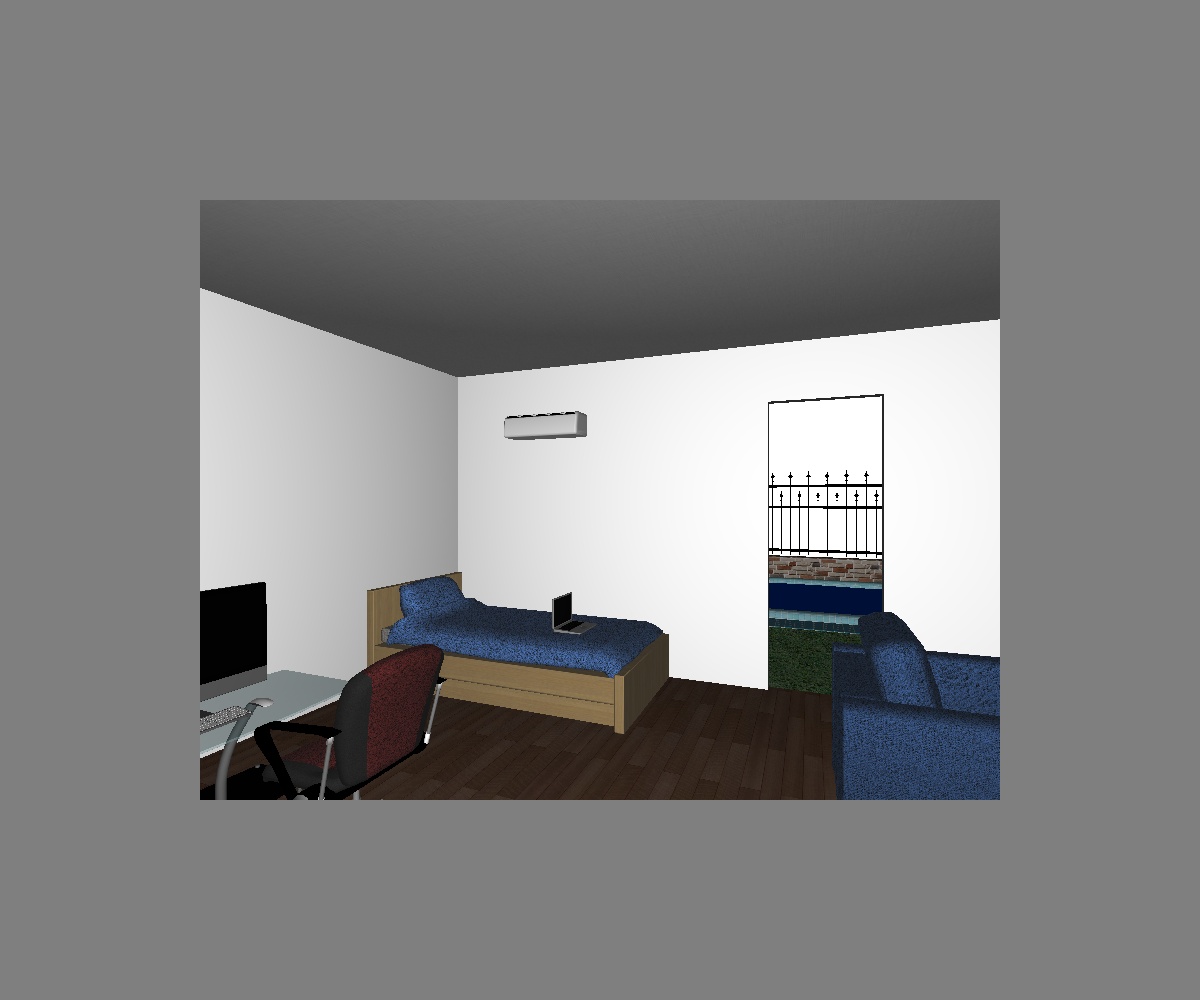}   	
    \end{tabular}
    \caption{Exemplar top-down maps of shortest paths (top row) and the corresponding observations at step 0, 3, 6, 9 and 12 (bottom five rows). }
    \label{fig:shortestpath}
\end{figure*}

For each spawned location in the environment, we compute a shortest path from the initial location to the target object. In Fig.~\ref{fig:shortestpath}, we show four examples. In the 2D top-down maps, the blue dot denote the target object; the red regions represent potential spawning locations of the agent; the green dots denote the selected spawning location; the blue curves are the shortest-path trajectories. The bottom five rows are agents' observations in each step.

\section{House Names in SUNCG}
In the text files (train.txt, val.txt, test.txt), we list all the house names for training, validation and test in \url{https://www.cc.gatech.edu/~jyang375/evr.html}.

\end{appendices}

\end{document}